\documentclass{article}
\usepackage[final,main]{neurips_2026}
\usepackage[utf8]{inputenc}
\usepackage[T1]{fontenc}
\usepackage{hyperref}
\usepackage{url}
\usepackage{booktabs}
\usepackage{amsfonts}
\usepackage{nicefrac}
\usepackage{microtype}
\usepackage{xcolor}
\usepackage{graphicx}
\usepackage{subcaption}
\usepackage{xspace}
\usepackage{placeins}
\usepackage{float}
\usepackage{amsmath,amssymb,amsthm,mathtools}
\usepackage{array}
\usepackage{tabularx}
\usepackage{multirow}
\usepackage{enumitem}
\usepackage{tikz}
\usepackage{wrapfig}
\theoremstyle{definition}
\newtheorem{remark}{Remark}

\usepackage[nodisplayskipstretch]{setspace}
\usepackage{changepage}
\usepackage{etoolbox}

\definecolor{panelgray}{RGB}{242,242,242}
\definecolor{modelgray}{RGB}{232,232,232}
\definecolor{rulegray}{RGB}{0,0,0}
\definecolor{predblue}{HTML}{1f77b4}
\definecolor{datagray}{HTML}{666666}
\definecolor{modelred}{HTML}{d62728}
\definecolor{measureyellow}{HTML}{837b00}
\colorlet{modelredtext}{modelred!90!black}

\AtBeginDocument{%
  \addtolength\abovedisplayskip{-0.55\baselineskip}%
  \addtolength\belowdisplayskip{-0.55\baselineskip}%
}
\makeatletter
\def\thm@space@setup{\thm@preskip=0.25em
\thm@postskip=0em}
\makeatother
\setlist{nolistsep,leftmargin=*}
\usepackage[all]{nowidow}

\newcommand{\Sref}[2][]{\hyperref[#2]{Sec.~\ref*{#2}#1}}
\newcommand{\Fref}[2][]{\hyperref[#2]{Fig.~\ref*{#2}#1}}
\newcommand{\Tref}[2][]{\hyperref[#2]{Tab.~\ref*{#2}#1}}
\newcommand{\Eref}[2][]{\hyperref[#2]{Eq.~\ref*{#2}#1}}
\newcommand{\Aref}[2][]{\hyperref[#2]{Appx.~\ref*{#2}#1}}
\newcommand{\Dref}[2][]{\hyperref[#2]{Def.~\ref*{#2}#1}}

\newcommand{\maybeincludegraphics}[2][]{%
    \IfFileExists{#2}{%
        \includegraphics[#1]{#2}%
    }{%
        \IfFileExists{figures/#2}{%
            \includegraphics[#1]{figures/#2}%
        }{%
            \fbox{\parbox[c][0.22\textheight][c]{0.95\linewidth}{Missing figure: \texttt{\detokenize{#2}}}}%
        }%
    }%
}

\newcommand{\TV}{\mathrm{TV}}

\newcommand{\bbR}{\mathbb{R}}
\newcommand{\Rec}{m} 

\theoremstyle{definition}
\newtheorem{definition}{Definition}

\hypersetup{%
    colorlinks=true,
    linkcolor={red!50!black},
    citecolor={green!80!black},
    urlcolor=magenta
}

\title{Mechanisms of Misgeneralization in\\Physical Sequence Modeling}

\makeatletter
\renewcommand{\@bottomtitlebar}{%
  \vskip 0.29in
  \vskip -\parskip
  \hrule height 1\p@
  \vskip -0.1in%
}
\renewcommand{\@noticestring}{%
  $^\dagger$ Equal advising. \quad \@trackname
}
\newcommand{\equaladvisingmark}{\raisebox{0.55ex}[0pt][0pt]{\scriptsize$\dagger$}}
\makeatother

\author{%
Kento Nishi$^{1,2,3,4,5,6}$
\quad
Raphael Tang$^{7}$
\quad
Karun Kumar$^{3}$
\\
\textbf{Core Francisco Park$^{4,5}$\equaladvisingmark}
\quad
\textbf{Hidenori Tanaka$^{4,5}$\equaladvisingmark}
\vspace{3pt}\\
{\small $^1$Harvard College \quad $^2$Harvard John A.\ Paulson School of Engineering and Applied Sciences}\\
{\small $^3$Comcast AI \quad $^4$CBS-NTT Program in Physics of Intelligence, Harvard University}\\
{\small $^5$Physics of Artificial Intelligence Laboratories, NTT Research, Inc. \quad $^6$MATS \quad $^7$Microsoft}\\
{\small Correspondence: \texttt{knishi@mit.edu}}
\vspace{-0.18in}
}

\begin{document}

\maketitle

\begin{abstract}
Generative sequence models are often trained to plan motion in physical domains, from robotics to mechanical simulations.
When constructing a dataset to train such a model, engineers may curate demonstrations to specify how trajectories should be distributed over a \emph{physical quantity} like travel distance or mechanical energy.
For example, a roboticist building a maze navigation agent might choose demonstrations whose travel distances cover a fixed range uniformly, hoping to constrain the agent's expected power usage.
We find that standard deep learning can violate this intent: each generated trajectory can seem plausible on its own, but the aggregate distribution over the physical quantity is wrong.
We call this failure \emph{physical misgeneralization}.
In this work, we identify and formalize the problem, develop and validate an account of its mechanism, and use that account to compare and design mitigations.
Using controlled synthetic tasks, we show that physical misgeneralization arises when local errors typical of the model class propagate through the physical measurement to shift the recovered distribution.
We estimate these errors with a \emph{data deviation kernel}, and we use it to predict which physical quantity values gain or lose mass
{under real diffusion sequence models trained on }both toy and applied trajectory generation tasks.
Finally, our mechanistic interpretation helps identify which mitigations are structurally promising.
We first compare interventions that modify the dataset and generative interface, then show that the mechanistic understanding we develop throughout this paper suggests a novel way to construct data representations informed by the kernel. We show that our proposed mitigation, which transforms the coordinate system
under which the generator learns and samples, effectively reduces quantity drift. %
\end{abstract}

\section{Introduction}
\label{sec:intro}
The advent of generative sequence models has reshaped how intelligent physical systems are built.
In domains like robotics and dynamical modeling, generative deep learning is now used to synthesize robot policies~\citep{Octo2024}, forecast multi-agent motion for autonomous driving~\citep{Jiang2023MotionDiffuser}, sample probabilistic weather futures~\citep{Price2025GenCast}, model molecular dynamics trajectories~\citep{Jing2024MDGen}, and even construct action-controllable environments~\citep{Bruce2024Genie}.
Many of these systems learn from fixed collections of demonstration trajectories, and so the dataset is often the most direct lever to specify what physical behavior the learned generator should reproduce.
This is especially clear in robot learning, where engineers select for state-action trajectories that contribute the most information to the policy~\citep{Hejna2025RobotDataCuration}, retarget demonstrations across object poses, scene layouts, robot arms, and dexterous hands~\citep{Mandlekar2023MimicGen,Jiang2025DexMimicGen}, and collect safe or unsafe rollouts for aircraft taxiing, aerial navigation, and dynamical systems~\citep{Chou2018LearningConstraints,Ciftci2024SafeGIL}.
Across these cases, the purpose of shaping the dataset is to make the learned model reproduce the chosen physical structure at generation time.

Consider, then, a roboticist building a maze navigation agent from demonstrations.
Their primary objective is to make the agent learn how to move from the start to the goal.
However, they may also care about other physical quantities: for instance, travel distance can affect power usage.
To constrain this behavior, the roboticist curates the demonstration paths so that travel distances follow a fixed distribution, hoping that the agent will reproduce that distribution when it generates new paths.
If we train a planning model on this collection, one might reasonably expect that samples from the planner would recover approximately the same mixture of travel distances.
However, in practice, this expectation fails.
A diffusion planner~\citep{Janner2022Diffuser} trained on D4RL Maze2D~\citep{Fu2020D4RL}, a standard maze-navigation benchmark, learns to solve the maze; yet, when we measure the travel distances of those generated paths, the distribution is shifted upward relative to that of the training data (\Fref{fig:natural}).
The model therefore learned enough about maze-like motion to produce recognizable trajectories, but it failed to preserve the physical mixture that the dataset was meant to specify.
We call this failure \emph{physical misgeneralization}.

\usetikzlibrary{arrows.meta,calc}
\pgfdeclarelayer{panelbg}
\pgfdeclarelayer{arrowlayer}
\pgfsetlayers{panelbg,arrowlayer,main}

\tikzset{
  flow/.style={
    draw=black,
    line width=1.8pt,
    line cap=round,
    line join=round,
    postaction={
      draw=white,
      opacity=0.435,
      line width=0.55pt,
      line cap=round,
      dash pattern=on 1.15pt off 3.1pt
    },
    -{Latex[length=2.6mm,width=2.05mm]}
  },
  panel/.style={draw=none,fill=panelgray,rounded corners=7pt},
  blank/.style={inner sep=0pt,outer sep=0pt,anchor=center},
  rowlabel/.style={
    rounded corners=2.4pt,
    inner xsep=2.0pt,
    inner ysep=1pt,
    font=\fontsize{4.0}{4.0}\sffamily\bfseries,
    text=white
  },
  measurelabel/.style={
    font=\fontsize{5.8}{6.4}\sffamily\bfseries, 
    text=measureyellow!100,
    inner sep=0pt,
    anchor=center
  },
  assetoutline/.style={
    draw=black,
    line width=0.55pt,
    fill=none
  },
  tilelabel/.style={
    rowlabel,
    anchor=south west
  }
}

\ifdefined\FigOneStandalone
\begin{figure}[H]
\small
\centering
\else
\begin{wrapfigure}{r}{0.5\textwidth}
\vspace{-2.7em}
\small
\centering
\fi

\vspace{0pt}
\centering
\hspace{1em}\scalebox{0.9}{
\begin{tikzpicture}[x=1cm,y=1cm]

\def\xData{0.54}
\def\xBranch{1.72}
\def\xLeft{2.22}
\def\xTraj{4.08}
\def\xLaneA{5.24}
\def\xLaneB{5.48}
\def\xLaneC{5.72}
\def\xRuler{5.58}
\def\yTop{2.80}
\def\yMid{1.85}
\def\yBot{0.90}
\def\yRuler{0.31}
\def\yMeasure{-0.09}
\def\yPlot{-1.75}
\def\xPlotRight{4.5}
\def\xPlotLaneA{5.04}
\def\xPlotLaneB{5.28}
\def\xPlotLaneC{5.52}
\def\yPlotTopHit{-1.78}
\def\yPlotMidHit{-1.54}
\def\yPlotBotHit{-1.30}
\def\dataS{1.54}
\def\mazeS{1.54}
\def\assetHalf{0.77}

\node[blank] (data) at (\xData,\yMid) {\includegraphics[width=\dataS cm]{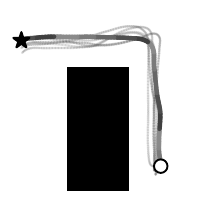}};

\path[fill=white] ($(\xLeft,\yTop)+(-\assetHalf,-\assetHalf)$) rectangle ($(\xLeft,\yTop)+(\assetHalf,\assetHalf)$);
\node[blank] (model) at (\xLeft,\yTop) {\includegraphics[width=1.05cm]{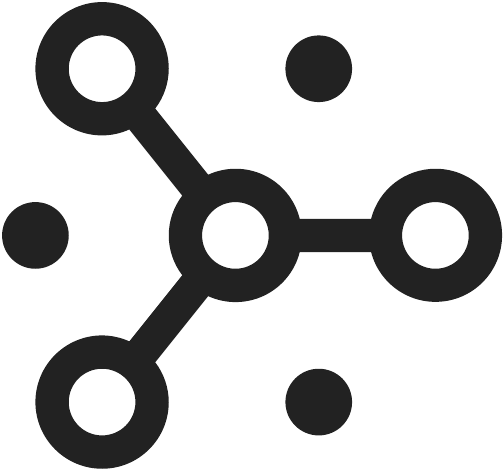}};

\node[blank] (kernel) at (\xLeft,\yBot) {\includegraphics[width=\mazeS cm]{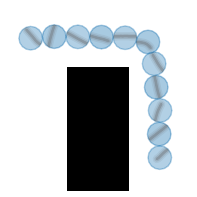}};

\node[blank] (trajModel) at (\xTraj,\yTop) {\includegraphics[width=\mazeS cm]{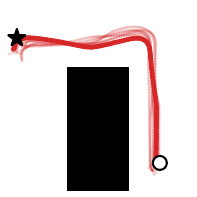}};
\node[blank] (trajKernel) at (\xTraj,\yBot) {\includegraphics[width=\mazeS cm]{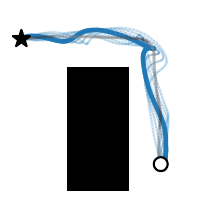}};

\node[blank] (curveplot) at (1.96,\yPlot) {\includegraphics[width=5.05cm]{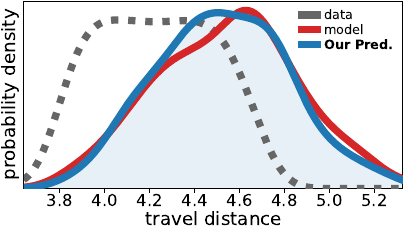}};

\coordinate (dataTopExit) at (data.north);
\coordinate (dataBotExit) at (data.south);
\begin{pgfonlayer}{arrowlayer}
\draw[flow,draw=predblue] (dataBotExit) -- (dataBotExit |- kernel.west) -- (kernel.west) -- (kernel.east) -- (trajKernel.west) -- (trajKernel.east) -- (\xPlotLaneA,\yBot) -- (\xPlotLaneA,\yPlotBotHit) -- (\xPlotRight,\yPlotBotHit);
\draw[flow,draw=modelred] (dataTopExit) -- (dataTopExit |- model.west) -- (model.west) -- (model.east) -- (trajModel.west) -- (trajModel.east) -- (\xPlotLaneC,\yTop) -- (\xPlotLaneC,\yPlotTopHit) -- (\xPlotRight,\yPlotTopHit);

\draw[flow,draw=datagray] (data.east) -- (\xPlotLaneB,\yMid) -- (\xPlotLaneB,\yPlotMidHit) -- (\xPlotRight,\yPlotMidHit);
\end{pgfonlayer}

\node[measurelabel] (measuretext) at (3,\yMeasure) {physical quantity measurement};
\draw[measureyellow!90,line width=0.45pt,densely dotted] (-0.22,\yMeasure) -- ($(measuretext.west)+(-0.05,0)$);
\draw[measureyellow!90,line width=0.45pt,densely dotted] ($(measuretext.east)+(0.025,0)$) -- (6.35,\yMeasure);

\draw[assetoutline] ($(data.center)+(-\assetHalf,-\assetHalf)$) rectangle ($(data.center)+(\assetHalf,\assetHalf)$);
\draw[assetoutline] ($(model.center)+(-\assetHalf,-\assetHalf)$) rectangle ($(model.center)+(\assetHalf,\assetHalf)$);
\draw[assetoutline] ($(kernel.center)+(-\assetHalf,-\assetHalf)$) rectangle ($(kernel.center)+(\assetHalf,\assetHalf)$);
\draw[assetoutline] ($(trajModel.center)+(-\assetHalf,-\assetHalf)$) rectangle ($(trajModel.center)+(\assetHalf,\assetHalf)$);
\draw[assetoutline] ($(trajKernel.center)+(-\assetHalf,-\assetHalf)$) rectangle ($(trajKernel.center)+(\assetHalf,\assetHalf)$);

\node[tilelabel,fill=datagray,draw=datagray] at ($(data.center)+(-\assetHalf+0.025,-\assetHalf+0.025)$) {dataset};
\node[tilelabel,fill=modelred,draw=modelred] at ($(model.center)+(-\assetHalf+0.025,-\assetHalf+0.025)$) {model};
\node[tilelabel,fill=predblue,draw=predblue] at ($(kernel.center)+(-\assetHalf+0.025,-\assetHalf+0.025)$) {data dev. kernel};

\node[blank] (ruler) at (\xRuler,\yRuler) {\includegraphics[width=1.62cm]{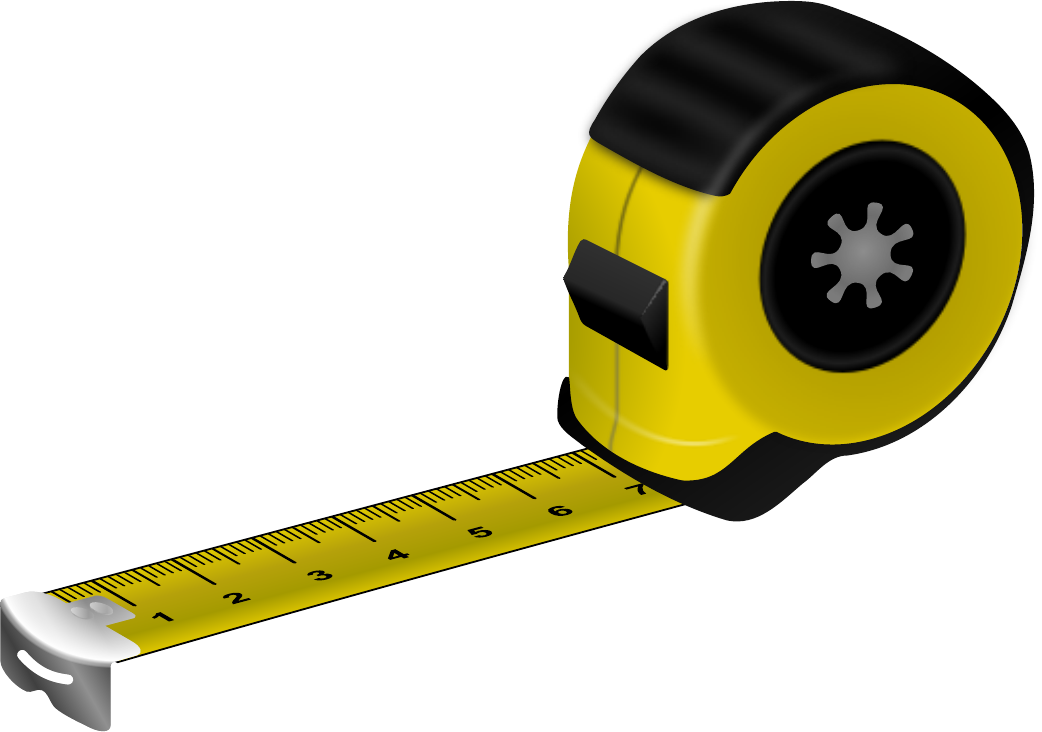}};

\end{tikzpicture}
}
\vspace{-0.5em}
\caption{\textbf{We identify the mechanism by which sequence models fail to match the distribution of a physically measured quantity.}
Imagine training an agent to navigate a maze, with a dataset curated so the distribution of travel distances falls in a safe range.
After training, the model can solve the maze, but its paths have longer travel distance than the ones in the {\color{datagray}\textbf{training data}}.
We unpack why: \textbf{\color{modelredtext}local errors typical of the model class} get propagated through the \textbf{\color{measureyellow}distance measurement} to cause drift.
By approximating model errors as a \emph{\textbf{\color{predblue}data deviation kernel}}, we can closely predict how the model shifts the distribution.}
\label{fig:natural}

\ifdefined\FigOneStandalone
\end{figure}
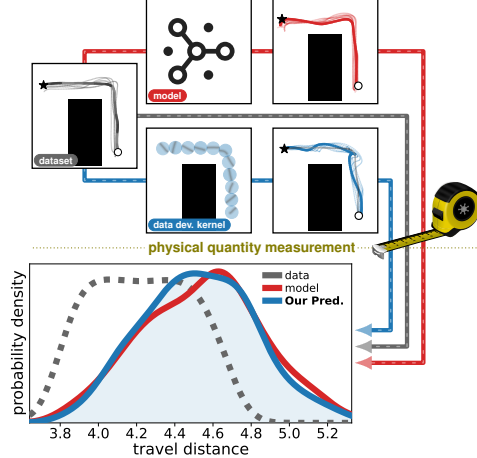
\else
\vspace{-1.5em}
\end{wrapfigure}
\fi

Why does this happen?
Peering into how the model is trained, travel distance does not explicitly appear as an optimization target for the diffusion model.
Consequently, we are asking the model to place approximately the right amount of probability in each measured distance bin, without telling it that travel distance is an evaluation criterion.
This demand is stronger than producing valid plans: the planner can introduce slight detours while moving through the maze, and those deviations may be small from the perspective of the trajectory-space loss, but they may still have large effects on the distances traveled.
The mechanism at play, then, seems to be that the marginal objective allows local trajectory deviations that preserve recognizable motion, but those same deviations convert into uneven transfer of probability mass along the measured physical quantity of interest.

As stated, this mechanistic interpretation is only conjectural.
To validate it, we need evidence that it correctly predicts empirical observations, \textit{i.e.}, that it can predict the movement of probability mass in real trained models.
Natural physical datasets rarely expose all pieces needed to carry out such tests, because the measured quantity, the relationships between quantities and trajectories, and the modeling method can all be deeply entangled.
Motivated by this, we construct synthetic tasks in which a known scalar quantity governs a family of trajectories.
For each task, we fix an intended prior, map quantities to sequences through an explicit trajectory-generating rule, and recover the quantity from generated trajectories through a shared measurement method.
We still train the generator unconditionally on trajectories, preserving the structure of the navigation vignette; but, because we know the structure of the data generation process by construction, these tasks let us isolate the mechanism from the confounds of natural physical datasets.
In both the synthetic setups and applications like 2D maze navigation and double-pendulum simulation, we show that emulating model-like errors with a \emph{data deviation kernel} and propagating its perturbations through the physical measurement helps us forecast which quantity values gain or lose mass, without using the trained model's residual directions. Our key contributions are as follows:
\begin{itemize}[noitemsep,topsep=-0.25em,leftmargin=*]
    \item {\textbf{We identify and formalize \emph{physical misgeneralization}}, a failure mode that arises in the context of physical sequence models trained under marginal objectives. We observe this failure with diffusion models, autoregressive Transformers, and VAEs. (\Sref{sec:intro}, \Sref{sec:related}, \Sref{sec:definition}, \mbox{
\Aref{sec:app:hypothesis}}\hskip0pt
})%

    \item \textbf{We develop a mechanistic interpretation and validate its predictions.} %
    Model deviations, estimable as a \emph{data deviation kernel}, propagate through the physical measurement to move probability across quantity values.~(\Sref{sec:mechanism})
    We design synthetic tasks with known physical quantities that let us control the priors, trajectory-generating rules, and physical quantity measurements to test the mechanism's predictions.~(\Sref{sec:synthetic-suite})
    We show our mechanism strongly predicts which quantities get over- or under-represented by diffusion sequence models in both synthetic and applied tasks.
    It also explains the motivating maze navigation case.~(\Sref{sec:experiments})

    \item \textbf{We use the mechanistic interpretation to compare and design mitigations.}
    We compare interventions at varying levels: the dataset composition, generative model interface, and the input-output representation.
    We then use the kernel concept to propose a data representation that reduces drift in practice~(\Sref{sec:mitigations}).
\end{itemize}
In summary, this paper establishes that physical misgeneralization is an anticipatable consequence of the interaction between local model-induced deviations and quantities that govern the physical world.

\section{Related Work}
\label{sec:related}

\textbf{Generative sequence models for physical trajectories.}
These models are popular in trajectory-based planning and forecasting for autonomous driving, weather, molecular dynamics, and environment simulation~\citep{Janner2022Diffuser,Ajay2023DecisionDiffuser,Chi2023DiffusionPolicy,Octo2024,Jiang2023MotionDiffuser,Price2025GenCast,Jing2024MDGen,Bruce2024Genie}.
Separately, several works have also proposed explicitly physics-aware architectures: Hamiltonian Neural Networks learn vector fields through a Hamiltonian and thereby preserve conservation laws via inductive biases~\citep{Greydanus2019HNN}; Lagrangian Neural Networks parameterize Lagrangians in settings where canonical momenta are unavailable~\citep{Cranmer2020LNN}.
Although hardcoding physical laws can be valuable, the free-form generative sequence modeling approach remains widely used because it affords greater flexibility.
We therefore focus on the most common marginal modeling paradigm, in which the generator is trained without baking the organizing physical structure into the model interface.

\textbf{Latent variables, aggregate posteriors, and inverse problems.}
The act of recovering quantities from data algebraically resembles computing the aggregate posterior in a variational autoencoder.
In VAEs, a prior and likelihood define a latent-variable model, a learned recognition model maps observations to an approximate posterior, and averaging that posterior over observations gives the aggregate posterior~\citep{Kingma2013,Blei2017}.
ELBO surgery analyzes how this average encoding distribution differs from the prior~\citep{Hoffman2016}; adversarial autoencoders and Wasserstein autoencoders then regularize the encoded distribution toward the prior~\citep{Makhzani2015,Tolstikhin2017}.
The same algebraic resemblance extends to inverse problems, where Bayesian formulations combine a prior and likelihood into a posterior over hidden quantities, while classical treatments of ill-posedness and identifiability show that small perturbations in observed data can change the recovered quantity~\citep{Tarantola2005InverseProblemTheory,KaipioSomersalo2005InverseProblems,TikhonovArsenin1977IllPosed,BellmanAstrom1970Identifiability}.
Nonetheless, previous literature does not study the case central to this paper, wherein a generative model is trained marginally on data, and the distribution of the physical quantity is recovered by measuring generated samples.

\textbf{Biases of marginal generative models.}
Literature on sampling bias explains how the distribution learned by a generative model can systematically differ from the ground-truth data distribution.
For diffusion models, this includes mode interpolation, signal leakage, selective underfitting of some regions, local recombination of training examples, and closeby detours~\citep{Aithal2024ModeInterpolation,Everaert2024SignalLeak,Song2025SelectiveUnderfitting,Kamb2024DiffusionCreativity,Zhao2023DecisionStacks}.
Autoregressive sequence models are prone to similar issues like exposure bias, since teacher-forced training conditions on data prefixes whereas free-running generation conditions on the model's own sampled history~\citep{Schmidt_2019,Bengio2015ScheduledSampling,Lamb2016ProfessorForcing,Ross2011DAgger}.
These works mainly focus on individual sample quality via metrics like likelihood, perplexity, FID, and other notions of predictive fit~\citep{Meister2021BeyondPerplexity, Naeem2020FidelityDiversity}; our work complements these works by interrogating whether the aggregate of those individual samples obeys the intended distribution over a measured physical quantity absent from the model interface.

\textbf{Synthetic data generation for mechanistic understanding.}
The interaction between generator error and measurement sensitivity is difficult to isolate in natural physical datasets alone.
As such, we construct synthetic tasks in the same spirit as past mechanistic studies which designed controlled tasks to expose structure that would have otherwise remained entangled.
Synthetic tasks have been used to advance the field's understanding of compositional generalization, reasoning, training dynamics, and internal representations~\citep{Chan2022DataDistributionalProperties,Li2023EmergentWorldRepresentations,Nanda2023ProgressMeasures,Reddy2023MechanisticBasis,Brinkmann2024SymbolicReasoning,Lubana2024Percolation,Yang2024ConceptLearningDynamics,Park2025InContextLearningRepresentations,Nishi2025RepresentationShattering}.
This is made possible by the fact that knowing how the data are constructed allows for intervening and ablating parts of the pipeline to pick apart which components produce the behavior in question.
For physical misgeneralization, this means imposing a prior over a physical quantity, defining a rule to generate a trajectory for each quantity value, and fixing a measurement rule to recover the quantity.
With these elements in place, we can synthetically generate data, train models on that data, vary parts of the procedure, and identify the mechanistic root cause that drives the mismatch.

\section{Problem Formalization}
\label{sec:definition}

\begin{wrapfigure}{r}{0.33\textwidth}
    \vspace{-5\baselineskip}
    \centering
    \includegraphics[width=\linewidth]{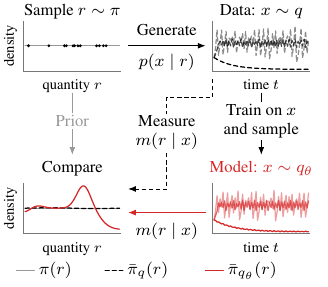}
    \captionsetup{font=footnotesize,skip=3pt}
    \caption{{\textbf{A schematic of the pipeline.}
    \emph{(1)} Sample $r\sim\pi$, then $x\sim p(\cdot\mid r)$.
    \emph{(2)}~Train $q_\theta$ on trajectories $x$.
    \emph{(3)}~Apply $\Rec$ to data and model samples.
    \emph{(4)}~Compare marginals $\bar{\pi}_q$ and $\bar{\pi}_{q_\theta}$ with $\pi$.
    }}
    \label{fig:formalization-pipeline}
    \vspace{-1.5\baselineskip}
\end{wrapfigure}
We now formalize the failure by separating all moving parts: namely, the physical quantity, measurement, trajectories, and model. We provide a visual aid to follow along in~\Fref{fig:formalization-pipeline}.

Let $r\in\mathcal R\subseteq\bbR$ denote the scalar quantity whose marginal distribution we measure, let $z$ collect the remaining variation, and let $x$ denote the resulting trajectory.
We use $z$ to index the within-quantity variation in $x$, \textit{i.e.}, how trajectories can still differ after we fix $r$.
For example, mechanical energy ($r$) does not uniquely identify double-pendulum motion ($x$): two pendulums can start from different positions with the same energy.
In this illustrative example, $z$ would describe the differing initial positions.

Suppose data are generated by first drawing $(r,z)$ from a joint distribution $\rho(r,z)$ and then drawing $x\sim p(x\mid r,z)$.
Since the training data supplied to the model contain only trajectories, the marginal distribution seen during training is
\begin{equation}
q(x) \;=\; \int p(x \mid r,z)\,\rho(r,z)\,dr\,dz.
\label{eq:trajectory-marginal}
\end{equation}
The target distribution over the physical quantity is the marginal
\begin{equation}
\pi(r) \;=\; \int \rho(r,z)\,dz.
\label{eq:quantity-marginal}
\end{equation}
Thus, although we measure a scalar marginal, the remaining variation can still be coupled to that quantity through the joint data-generating process.
For the scalar comparison used in our experiments, we integrate this variation into an effective conditional trajectory family,
\begin{equation}
p(x\mid r)\;=\;\int p(x\mid r,z)\,\rho(z\mid r)\,dz,
\label{eq:nuisance-absorb}
\end{equation}
so that equivalently $q(x)=\int p(x\mid r)\,\pi(r)\,dr$.
In this reduced form, preserving the intended mixture means recovering $\pi(r)$ after the remaining variation has been averaged into $p(x\mid r)$.
When more quantities are relevant, we can use a vector-valued $\vec{r}$ or a projection of the quantities.

\begin{remark}[Prior absorption]
We can also absorb the shape of the prior into the scalar coordinate.
For any scalar prior $\pi$, let $T:[0,1]\to\mathcal R$ be a transport map with $T_\#\mathrm{Unif}([0,1])=\pi$; for a continuous prior, this is the usual quantile map $T(u)=F_\pi^{-1}(u)$.
Writing $r=T(u)$ with $u\sim\mathrm{Unif}([0,1])$ gives
\begin{equation}
q(x)
=\int p(x\mid r)\,\pi(r)\,dr
=\int_0^1 p\!\left(x\mid T(u)\right)\,du,
\label{eq:uniform-absorb}
\end{equation}
so we can absorb any non-uniform prior into a reparameterized map from quantity values to trajectories while leaving the marginal training distribution unchanged.
Thus, we can generally use a uniform prior over the reparameterized quantity for convenience, composing the data family with $T$.\hfill$\triangle$
\end{remark}

We write $q_\theta$ for the trajectory distribution learned from samples of $q$, where $\theta$ parameterizes the model.
Because both $q$ and $q_\theta$ are trajectory distributions, we define their mixtures over the physical quantity by applying a measurement rule to trajectories.
We write this rule as $\Rec(r\mid x)$; it may return a point value or a distribution over candidate quantity values. We adopt the latter formulation of $\Rec(r\mid x)$ in our synthetic experiments, because an imperfect trajectory may admit several comparably good estimates.
Concretely, we compare a trajectory with reference trajectories generated at candidate quantity values, assigning higher probability to closer matches~(\Eref{eq:posterior}, \Aref{sec:app:synthetic-recovery}).
We apply the same $\Rec$ to data and model samples.
Fixing $\Rec$ in this way allows us to isolate changes in the trajectory distribution from changes in the measurement.
Under these conventions, for any source distribution $S$ over trajectories, the induced marginal over the physical quantity is therefore
\begin{equation}
\bar{\pi}_{S}(r) \;=\; \mathbb{E}_{x\sim S}\big[\Rec(r\mid x)\big].
\label{eq:pullback}
\end{equation}

\begin{definition}[Quantity drift]
\label{def:drift}
Let $\bar{\pi}_{\mathrm{data}} \coloneqq \bar{\pi}_{q}$ and $\bar{\pi}_{\mathrm{model}} \coloneqq \bar{\pi}_{q_\theta}$ under a shared method for measuring the physical quantity.
For any distributional distance or divergence metric $d$, the $d$-measured quantity drift of $q_\theta$ is
\begin{equation}
    \mathrm{Drift}_d(q_\theta) \coloneqq d(\bar{\pi}_{\mathrm{model}}, \pi).
\end{equation}
We interpret this value against the data baseline $d(\bar{\pi}_{\mathrm{data}},\pi)$.
We say the model has substantial drift under $d$ when $\mathrm{Drift}_d(q_\theta) \gg d(\bar{\pi}_{\mathrm{data}},\pi)$. \hfill$\triangle$
\end{definition}
In practice, we mainly report the drift using total variation,
\begin{align}
	\TV(\bar{\pi}_{q_\theta}, \pi)
	&= \frac{1}{2}\int_{\mathbb{R}} \bigl| \bar{\pi}_{q_\theta}(r) - \pi(r) \bigr| dr.
	\end{align}
We use total variation because it directly measures how much probability mass must move between quantity values.
Alternative choices, such as Jensen--Shannon divergence, Hellinger distance, Wasserstein--1, and KL divergence under compatible supports, also suffice.

\section{Predicting Quantity Drift with the Data Deviation Kernel}
\label{sec:mechanism}

To predict how the distribution changes, we further need a signed redistribution relative to the recovered data marginal.
We write this signed drift as
\begin{equation}
\tau_\theta(r)\coloneqq \bar{\pi}_{q_\theta}(r)-\bar{\pi}_{q}(r),
\label{eq:signed-drift}
\end{equation}
so that positive values correspond to quantity values over-represented by the model, while negative values correspond to quantity values depleted by generation.
Since both recovered marginals are obtained by applying the same measurement rule to trajectory distributions, expanding \Eref{eq:signed-drift} gives
\begin{equation}
\tau_\theta(r)
=
\int \bigl(q_\theta(x)-q(x)\bigr)\Rec(r\mid x)\,dx.
\label{eq:transport}
\end{equation}
Under this expansion, a trajectory-space error contributes mass to the quantity value recovered for the corresponding trajectory.
When the model places excess probability on trajectories recovered as $r$, it increases $\tau_\theta(r)$; conversely, missing probability decreases $\tau_\theta(r)$.
That is to say, the effect of a trajectory-space discrepancy depends on where that discrepancy lies relative to the measurement rule.

If we knew the signed model error $q_\theta-q$, we could compute the drift directly from \Eref{eq:transport}.
Our goal, though, is to anticipate this signed movement without estimating $q_\theta$, the hypothetical model, as a full distribution.
To do this, we need a local surrogate for the signed error induced when a given architecture is trained on a given dataset and objective.
\begin{definition}[Data deviation kernel]
\label{def:kernel}
The conditional distribution
$
D_{\mathrm{arch}}^{\sigma}(\delta\mid x)
$
over deviations $\delta$ around $x$ summarizes how a specific model architecture family might readily disperse the probability mass of each data point. The scalar $\sigma$ controls the strength of the deviations applied by the kernel. \hfill$\triangle$
\end{definition}
Importantly, $\sigma$ is a free strength parameter; when making a prediction, one can—but is not required to—choose $\sigma$ so the resulting deviation magnitude roughly matches a target total error scale, such as the loss level of the model quality one wishes to forecast.
Consider the extremes: a perfect model and a randomly initialized model with the same architecture.
The former reproduces the data marginal, and hence its recovered quantity distribution, exactly; the latter need not.
These two cannot share one useful predicted quantity distribution.
Calibration serves to account for this difference in model fitness: when we use the model's data-space error to set the deviation magnitudes, we make a \emph{model-calibrated comparison}~(\Sref{sec:synthetic-suite}, \Aref{sec:app:scale}).
Note that it is also possible in principle to vary $\sigma$ without any information about the target model to obtain a \emph{model-free relationship} between induced data-space error and quantity drift.
In this sense, the kernel's functional form remains agnostic to the trained model and can predict drift before training, a priori, with no reliance on residual directions, generated samples, the recovered marginal over the physical quantity, or a separately trained model.

\begin{remark}[Kernel for diffusion]
Recall from \Sref{sec:related} that diffusion models are known to distort the learned distribution in data-space through mode interpolation, signal leakage, selective underfitting, local recombination, and closeby detours~\citep{Aithal2024ModeInterpolation,Everaert2024SignalLeak,Song2025SelectiveUnderfitting,Kamb2024DiffusionCreativity, Zhao2023DecisionStacks}.
These results together suggest that, around a data trajectory $x$, trained models move probability towards nearby pieces of trajectories inasmuch as the model or sampler readily confuses those fragments.
To reflect this, we instantiate a simple local kernel: for each local state or window, we replace the original piece with a similar fragment drawn from the data using Gaussian distance weights.
This definition is derived from established diffusion phenomena, albeit as a tractable approximation. Specifically, local recombination~\citep{Kamb2024DiffusionCreativity} motivates local replacement, while mode interpolation~\citep{Aithal2024ModeInterpolation} and selective underfitting~\cite{Song2025SelectiveUnderfitting} motivate using empirically available swaps drawn from the training data. To allow perturbations beyond the observed fragments to ``make up'' for variation not already supplied by those fragments, we also add Gaussian noise~(details in \Aref{sec:app:kernel-impl}).
Intuitively, this kernel swaps small pieces of each trajectory with fragments from other trajectories that look similar, with noise accounting for any remaining local perturbation. \hfill$\triangle$
\end{remark}

With $D_{\mathrm{arch}}^{\sigma}$ specified, we predict drift by asking which quantity values are recovered after kernel perturbations.
For a trajectory drawn from $p(x\mid r)$, we sample a deviation from $D_{\mathrm{arch}}^{\sigma}(\cdot\mid x)$ and measure the perturbed trajectory $x+\delta$.
Averaging over both draws gives the quantity-transport kernel
\begin{equation}
K_{\mathrm{dev}}(r'\mid r)
\coloneqq
\mathbb{E}_{x\sim p(x\mid r)}
\mathbb{E}_{\delta\sim D_{\mathrm{arch}}^{\sigma}(\cdot\mid x)}
\left[\Rec(r'\mid x+\delta)\right].
\label{eq:addk-kernel}
\end{equation}
In words, $K_{\mathrm{dev}}(r'\mid r)$ is the probability that a trajectory associated with $r$ is recovered as $r'$ after a deviation.
We then apply this transition to the intended prior to obtain the predicted $r$ marginal
\begin{equation}
\bar{\pi}_{\mathrm{dev}}(r')
=
\int K_{\mathrm{dev}}(r'\mid r)\pi(r)\,dr.
\label{eq:predicted-marginal}
\end{equation}
Thus, with just a data deviation kernel and the recovery rule used to measure trajectories, $\bar{\pi}_{\mathrm{dev}}$ predicts which quantity values the generative model may over- or under-represent.

For a concrete example for when drift can occur at all, suppose the quantity takes only two values $r_1,r_2$ with intended probability $1/2$ each, and the measurement recovers these values exactly on the data.
Let $K_{ij}$ be the probability that a trajectory generated from $r_i$ is measured as $r_j$ after a deviation.
The recovered probability at $r_2$ is then $\tfrac12 K_{12}+\tfrac12 K_{22}=\tfrac12+\tfrac12(K_{12}-K_{21})$.
For the recovered marginal to match the uniform prior, each quantity must receive as much probability as it loses.
If trajectories from $r_1$ are more often measured as $r_2$ after a deviation than the reverse, probability accumulates at $r_2$.
Thus, local errors in $x$-space can induce asymmetric movement in $r$-space.

\section{Experiments: Validating the Mechanistic Interpretation}

\subsection{Defining the Data Generation Processes}
\label{sec:synthetic-suite}
\paragraph{Controlled synthetic tasks.} For the predictive calculation to explain physical misgeneralization, it must anticipate the drift of trained models.
We first test this with synthetic constructions because they give us direct control over the quantity that generates each trajectory, while leaving the model in the same marginal training setup as explained in the earlier vignettes.
For each task, we draw a scalar quantity $r\sim\mathrm{Unif}(\mathcal R)$, set $x_0=0.25$, generate a length-$H$ trajectory $x=g_{\mathcal S}(r)$, and train a marginal 1D U-Net diffusion model on trajectories alone.
The intended mixture over $r$ therefore touches training only through the trajectories.
After generation, we infer the model's induced mixture over $r$ from samples in the same way that we infer it from ground-truth data.
Throughout the synthetic suite, we use sequence horizon length $H=64$ and consider three one-dimensional physical sequence families.
We start with the sinusoid family
\begin{equation}
x_t=\tfrac12\left(\sin(2\pi t/r+\phi_0)+1\right),\qquad r\in[32,128],
\end{equation}
where $\phi_0=-\pi/6$ and $r$ is the period in timesteps.
We use this as a simple learnable baseline because increasing $r$ stretches the oscillation over time such that nearby periods produce nearby sequences over the horizon~(proof in \Aref{sec:app:lyapunov-sinusoid}).
We then use the tent map
\begin{equation}
x_{t+1}=r\min\{x_t,1-x_t\},\qquad r\in[0,2],
\end{equation}
which is piecewise linear and has slope magnitude $r$ at every differentiable state $x_t\neq 1/2$.
Solving the Lyapunov exponent yields $\lambda_H(r)=\log r$ (proof in \Aref{sec:app:lyapunov-tent}), meaning that sequences are non-chaotic for $r \leq 1$, while the exponential growth rate of local state perturbations 
increases logarithmically as $r$ approaches $2$.
Finally, we use the logistic map,
\begin{equation}
x_{t+1}=r x_t(1-x_t),\qquad r\in[0,4],
\end{equation}
as a sanity check. 
This is a classical nonlinear population dynamics model in which $r$ controls the growth rate.
The logistic map is challenging to predict because its Lyapunov exponent depends on realized orbits (proof in \Aref{sec:app:lyapunov-logistic}); we include it here to test whether our mechanism generalizes beyond the tent map.
For these iterated maps, we discretize intermediate states to $1024$ bins and use the same numerical recurrence whenever we generate data or evaluate the prediction.
Note that we deliberately construct these tasks to avoid variation over $z$, by fixing the phase for sinusoid and the initial condition for tent and logistic.
This lets us isolate drift over $r$ from variation over $z$.

\paragraph{Training, quantity recovery, and prediction details.}
For each model, we train on $200{,}000$ trajectories with a 1D U-Net diffusion backbone (base width $32$, kernel size $3$) and sample with $256$ DDPM steps.
For plotting, we use $25{,}000$ ground-truth trajectories and $10{,}000$ generated trajectories.
To recover the mixture over $r$, we compare each trajectory to reference trajectories on a grid of $2^{14}=16{,}384$ quantity values~(details in \Aref{sec:app:synthetic-recovery}
) and average the resulting recovery posteriors over trajectories.
For example, tent map trajectories for $r=0.80$ and $r=0.81$ rapidly decay toward zero, so an imperfect generation between them may receive comparable likelihood under either value.

To compare our prediction against each trained diffusion model, we choose the scale $\sigma$ so the overall scalar data-space deviation induced by the kernel matches the model's overall scalar data-space error.
We measure this error directly after generation as the aggregate root-mean-square residual between model samples and the data trajectories associated with their recovered quantity values, and use the variation in per-trajectory error magnitudes over the recovered quantity to set relative local kernel scales~(\Aref{sec:app:scale}). 
We also repeat each experiment with $3$ seeds to account for variance~(\Aref{sec:app:statistics}).
For additional details, see~\Aref{sec:app:readout}; for visualizations of data and generated trajectories, see~\Aref{sec:app:visualizations}.

\paragraph{Applied tasks: double-pendulum simulation and Maze2D navigation.} 
For double-pendulum, we generate position trajectories from the ground-truth equations, curate the data so mechanical energy is uniform over the range $[5,40]$, and train the diffusion model only on the two angles' coordinates.
We recover the quantity by estimating angular velocities from finite differences and measuring the median shifted energy along the rollout~(\Aref{sec:app:pendulum}).
For Maze2D, we train on fixed-horizon movement segments extracted from D4RL U-maze sparse-v1, curated to have a uniform distribution over path length over the range $[3.80,4.65]$.
After filtering wall collisions, we recover the quantity by measuring total path length from the generated position sequence~(\Aref{sec:app:maze2d}).

\begin{figure}[t]
    \centering
    \maybeincludegraphics[width=0.98\linewidth]{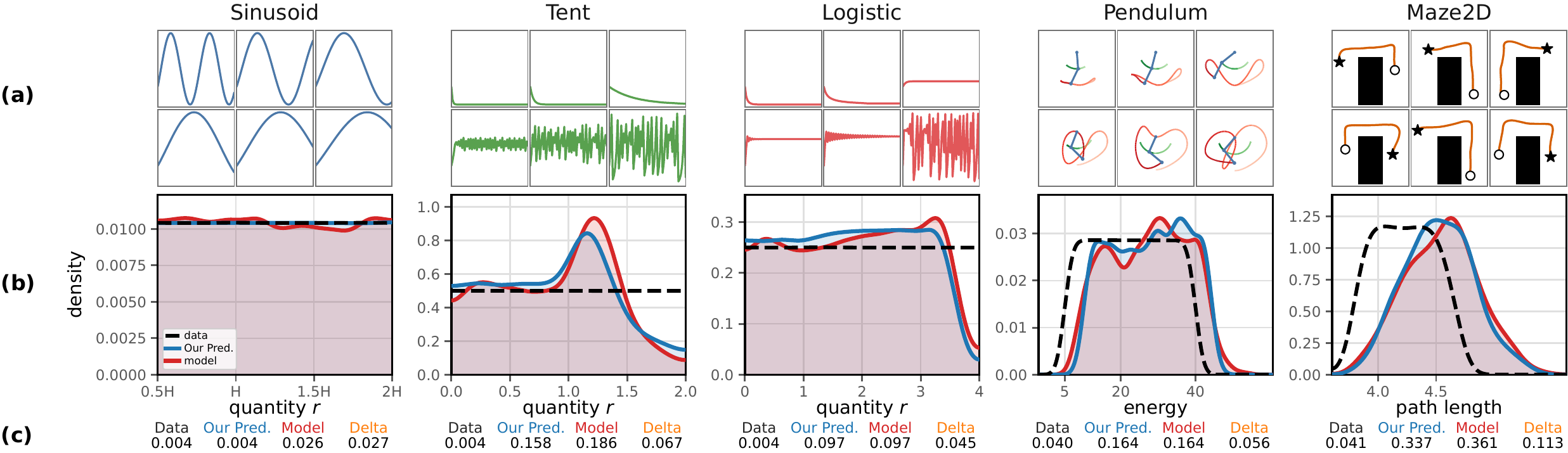}
    \vspace{-0.5em}
\caption{\textbf{Trained models closely replicate the mechanism's predicted physical quantity drift.}
    \emph{(a)}~Representative visualizations of trajectories in each dataset.
    \emph{(b)}~The mechanism (blue) predicts that the sinusoid curve will remain nearly flat like the intended prior, whereas for tent and logistic, it forecasts excess mass at intermediate $r$ and depleted mass near the upper end of the range.
    For double-pendulum, it forecasts a lateral and bumpy translation towards higher mechanical energy; for Maze2D, it suggests a tightened and shifted distribution of longer path lengths.
    The actual trained model (red) closely replicates our predictions, with little movement for sinusoid, mid-range gain followed by high-$r$ deficits for tent and logistic maps, and the expected drifts and shapes for double-pendulum and Maze2D.
    \emph{(c)}~The $\TV$ values compare distances from the prior, and the predictions' delta to the models. The small delta values indicate that our predictions forecast trained models well.}
    \label{fig:synthetic-kernel-pred-curves}
    \label{fig:synthetic-kernel-pred-tv}
    \label{fig:synthetic-kernel-pred}
    \vspace{-1em}
\end{figure}

\subsection{Results: Comparing the Prior, Data, Mechanistic Predictions, and Trained Model}
\label{sec:experiments}
In \Fref{fig:synthetic-kernel-pred}, we compare the distributions over quantities for the intended prior, the ground-truth data, the mechanism's predictions, and the model.
Immediately, we can see that the mechanism forecasts little movement for the sinusoid, a large mid-range rise and high-$r$ depletion for the tent map, a broader high-$r$ rise and drop for the logistic map, an upward and unstable energy shift in double-pendulum, and a reshaped and upward shift of path length in Maze2D.
Then, we inspect the distributions implied by the trained models: their curves closely track our predictions, with small $\TV$ deltas to the predicted curve in each case.
The sinusoid model remains close to the intended mixture, whereas the tent map model over-represents intermediate values around $r\in[1.1,1.4]$ and under-represents high values above roughly $r=1.5$; similarly, the logistic map places excess mass around $r\in[3.2,3.6]$ and depleted mass near the upper end of the range.
Further, the double-pendulum model matches our predicted energy distribution well, even reproducing where the intermediate plateau was transformed into minute peaks and troughs. 
The Maze2D planner also matches our prediction of a tighter, more triangular, and right-shifted distribution, with its peak aligning with that of the prediction.

We also find that the predicted profile is highly robust to changes in $\sigma$ (see \Aref{sec:app:scale}). In net, this agreement supports the mechanistic interpretation: local model-induced deviations, passed through the physical measurement, can help anticipate which quantity values gain or lose mass.

{For model-free predictions via scales selected without calibrating to information from trained models, comparisons across training checkpoints, and additional validations of the kernel, see \Aref{sec:app:checkpoint-predictions}--\ref{sec:app:kernel-validation}.}
Separately, for evidence that residuals between prediction and model are not systematically concentrated in regions with high Lyapunov exponents across tent and logistic maps, see~\Aref{sec:app:lyapunov-residual}.

\begin{figure}[t]
    \centering
    \maybeincludegraphics[width=0.98\linewidth]{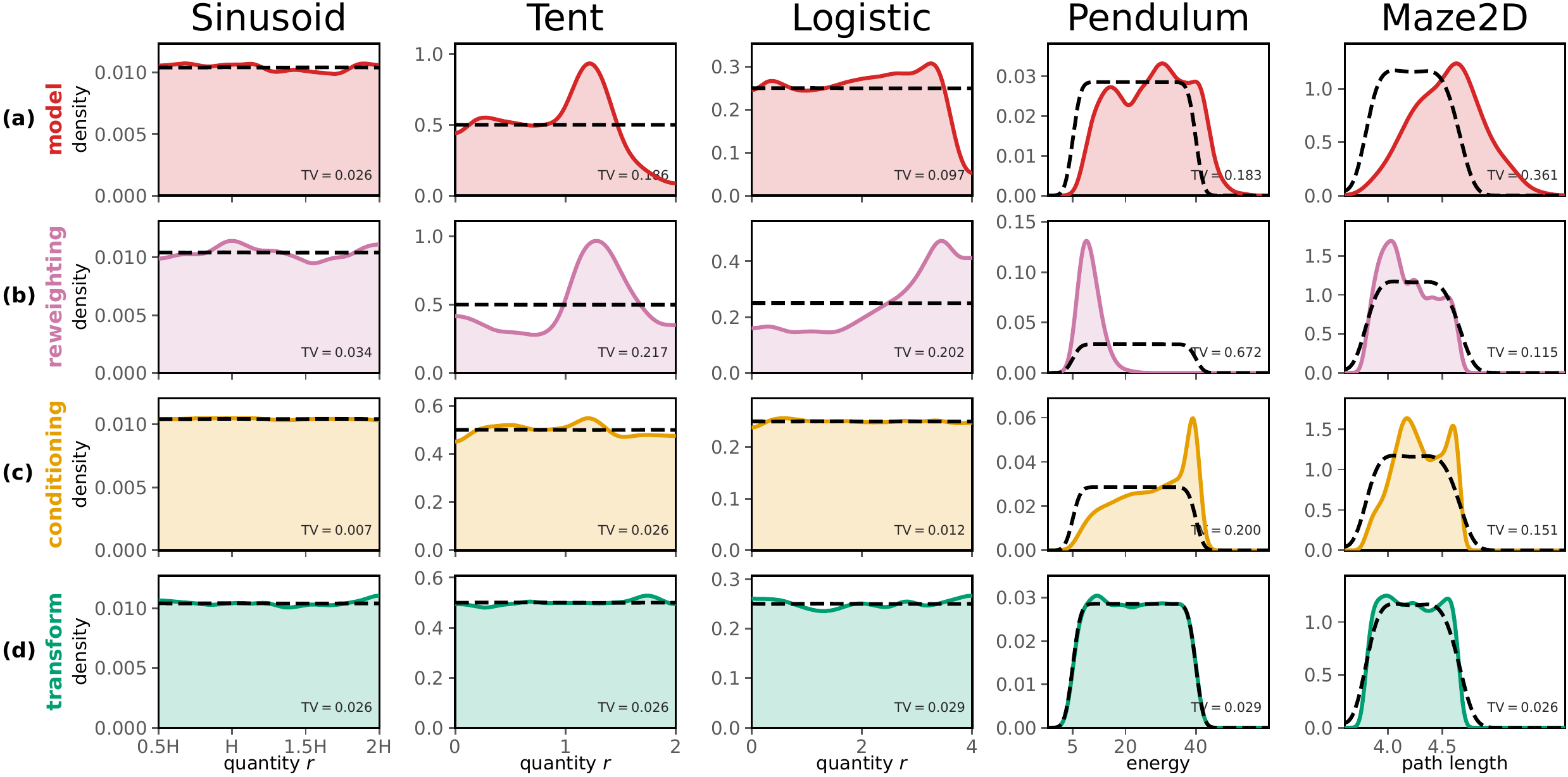}
    \vspace{-0.6em}
\caption{\textbf{Mechanism-informed interventions can reduce drift.}
    One can attempt to reduce drift in three distinct ways: \emph{(b)} rebalancing the dataset, \emph{(c)} modeling conditionally, and \emph{(d)} transforming the input-output data representation. The strongest and most consistent correction comes from using the kernel to derive a coordinate transformation that balances mass transfer between quantity values.}
    \label{fig:synthetic-generation-pipeline}
    \label{fig:physical-generation-pipeline}
    \vspace{-1em}
\end{figure}
\section{Mitigating Physical Misgeneralization}
\label{sec:mitigations}

We have thus far focused on predicting how probability moves; we now ask what can be done to reduce that movement.
Importantly, before trying mitigations, we took care to ablate and rule out many simpler potential causes in sampling, recovery, architecture, seeds, and more (\Aref{sec:app:hypothesis}).

\paragraph{Modifying the dataset.} Among the categories of possible interventions, perhaps the most immediate is one that modifies what the generator is trained on.
That is, one can re-curate the dataset via over- or under-sampling, filtering, stratification, weighted losses, and synthetic data generation to counteract the model's preferences over $r$.
Conveniently, for our experimental setups described in \Sref{sec:experiments}, these methods effectively all reduce to reweighted sampling, in which we adjust the probability of showing each trajectory to the model.
If dataset rebalancing is to be effective, re-training the model with data sampled by the inverse of the recovered quantity marginal of the initial model should noticeably reduce the drift.
However, when we apply this dataset correction~(details in~\Aref{sec:app:inverse}), we find that it is ineffective and sometimes counterproductive (\Fref[b]{fig:synthetic-generation-pipeline}).
The poor performance of the dataset-level rebalancing follows naturally from the mechanistic picture developed earlier: although it changes how often different quantity values appear in the training set, it does not change the fact that the generator is trained marginally on trajectories, and therefore does not prevent readily expressed local errors from carrying probability mass across quantity values.
Moreover, reweighting one measured quantity can silently change other structure in the data.
In the chaotic maps, for example, different regimes of $r$ induce different trajectory statistics under the recurrence; therefore, increasing the sampling rate of one part of the quantity range also changes the distribution of states shown to the generator.
The same concern becomes more severe in applied physical datasets, where path length, start and goal endpoints, environment geometry, energy, and other physical quantities need not vary independently.

\paragraph{Modifying the generative interface.}
One might attempt a more aggressive intervention that directly provides the desired quantity to the generator.
Conditional modeling and guided sampling are natural examples of this category; fortunately, classifier-free guidance~\citep{HoSalimans2021ClassifierFree} unifies these ideas, and we can use it to condition diffusion models
and test whether they can be steered towards requested values of $r$ picked according to the intended prior~(details in \Aref{sec:app:conditional}).
We begin by testing this idea in the synthetic suite, mirroring our analysis of the dataset-level interventions and prior mechanistic studies that use fully controlled constructions to isolate an effect, then upgrade to richer but still structurally interpretable settings to test its scope.
Conditional modeling passes the first minimum test~(\Fref[c]{fig:synthetic-generation-pipeline}): the trajectory-generating rule is evidently simple enough for the model to largely follow once $r$ is supplied.
However, success in this controlled setting does not imply success in the richer settings: in double-pendulum and Maze2D, the occupied range of energy and path length is partially repaired, but the recovered distributions still contain large density spikes~(\Fref[c]{fig:synthetic-generation-pipeline}).
This happens because conditioning retains two weaknesses of the dataset-level intervention: it does not compensate for local trajectory-space errors that move probability,
and it does not specify the joint distribution between the requested quantity and the remaining structure needed to realize it.
For double-pendulum, the first issue is especially visible because low energy requires smooth position trajectories whose finite-difference velocities remain small, so small position-space errors recover as excess energy.
For Maze2D, the second issue is especially visible because the requested path length must be jointly feasible with the start point, endpoint, maze geometry, and horizon, since a path that is too short cannot reach the goal.
Thus, modifying the generative interface can reduce drift when the requested quantity indexes a relatively simple trajectory family, but its utility diminishes when the quantity must be jointly realized with other physical structure.

{
\paragraph{Transforming the data representation.}
The mechanistic understanding developed throughout this paper suggests another category of intervention, one that transforms the coordinate system under which the generator learns and samples.
Applying domain knowledge to choose input-output representations is standard practice: spectrograms for audio, one-hot tokenization for language, and learned encoders for vision are common.
Now, with mechanistic knowledge of the model's expected error geometry, we can use the kernel to construct even more favorable data representations.
Concretely, since physical misgeneralization arises when local neighborhoods around trajectories are imbalanced in the density of recovered quantity values, we can reduce drift by changing the geometry in which those neighborhoods are formed.
We first construct an unlabeled Latin-hypercube support in the input shape expected by the diffusion model; these codes define the coordinate system in which the model will see the data.
We then apply the kernel in this code space to identify which code neighborhoods are likely to exchange probability mass.
Using these neighborhoods, we can pair code points and training trajectories such that local decoding neighborhoods recover a balanced mixture over quantity values, while preserving the proportion of training trajectories assigned to each quantity.
Then, we train an unconditional diffusion model with the codes as the model-facing data representation.
At generation time, the model samples codes, after which they are pulled back through the code mapping and evaluated with the usual recovery rule.
Here, the generator never receives $r$; therefore, it avoids introducing new sources of information into the interface as does conditional modeling.
\Fref[d]{fig:synthetic-generation-pipeline} confirms that transforming the model-facing data representation consistently gives strong corrections, causing far less drift than comparable methods across all the synthetic and applied systems.
Furthermore, this preserves desiderata like sample quality and diversity well~(see \Aref{sec:app:coordinate} for details and checks).
As a caveat, this strategy may not always be applicable, \textit{i.e.}, when a particular representation choice is non-negotiable.
For instance, an audio application may require a model that works in waveforms rather than spectrograms; analogous mandates could exist for physical sequence modeling.
Still, the core point of this strong result is that the kernel can be extended towards novel and effective mitigations, more of which we hope future works will explore.
}

\section{Limitations and Broader Implications}
\label{sec:limitations}

\paragraph{Kernels crafted for diffusion models.}
Our choice to focus on diffusion-based physical sequence models follows from both the empirical vignettes that motivate the paper and the surrounding literature: diffusion models are the de facto architecture in these physical modeling settings, and recent work has already begun to describe how learned diffusion samplers move probability mass in data space.
This gave us a principled starting point to implement a data deviation kernel that captures the essence of how diffusion models express errors around data.
However, the kernel remains a coarse operationalization of these tendencies, compressing architecture, optimization, finite data, and sampling effects into a tractable local movement model.
This explains why the predictions in \Fref{fig:synthetic-kernel-pred} capture the drift patterns strongly, even predicting minor peaks and troughs, but are not pixel-perfect.
Likewise, the model-facing data representation mitigation strategy in \Sref{sec:mitigations} targets neighborhoods implied by the approximate kernel rather than the true sampler itself.

The specificity to diffusion also limits the scope of the predictive account; indeed, in \Aref{sec:app:hypothesis}, we show that autoregressive Transformers, VAEs, and MLPs trained on the tent map task also misgeneralize the distribution over the recovered quantity, but we do not make predictions for them because we do not yet define appropriate kernels for these architectures. We encourage future works along these lines: if we can come to better understand how other generative model families err in data space, those characterizations can inform the design of data deviation kernels.
Once those kernels are specified, the mechanism developed in this paper can again be used to anticipate which quantity values gain or lose mass and to guide mitigation strategies of the kinds studied in \Sref{sec:mitigations}.
It is also possible that, given the effectiveness of the transform mitigation with our approximate kernel, the mitigation could be extended to Transformers and VAEs without a fully specified kernel; a coarse estimate of which local neighborhoods exchange probability may suffice to construct an effective coordinate transform.

\paragraph{Scope beyond physical sequence models.}
Intriguingly, our preliminary results with a text-to-speech model suggest that the phenomenon may also arise in settings that are not usually framed as physical sequence modeling.
In \Aref{sec:app:tts-vignette}, we show an example in which the speaking rate of audio generated by a Tacotron~2--HiFi-GAN pipeline~\citep{Shen2018Tacotron2,Kong2020HiFiGAN,Ravanelli2021SpeechBrain} shifts upward relative to the training corpus.
Although this mismatch might be an instance of physical misgeneralization, extending our current framework to explain the text-to-speech pipeline precisely would require composing kernels across multiple learned stages.
Concretely, an acoustic model first produces an intermediate representation, and a separate vocoder maps that representation into waveforms; since speaking rate is measured only in waveform space, the kernel must describe how errors of the acoustic model and vocoder interact with one another.
Future work should develop data deviation kernels for such composed pipelines, where each stage can reshape probability mass before the final measurement is applied.

\paragraph{Algorithmic bias and fair representation.}
This failure mode has natural analogues to fairness: a generative model may produce individually plausible samples while systematically changing how different groups, styles, or behaviors are represented in aggregate.
This becomes especially problematic when the shifted quantity corresponds to a protected or socially salient attribute, such that already underrepresented populations become further misrepresented.
Reducing this kind of misalignment a priori is difficult when we do not know which axes of variation are consequential; however, when there are attributes or behaviors that practitioners have reason to care about, our results support evaluating and tuning generative models along those axes explicitly, in the spirit of disaggregated evaluations~\citep{Barocas2021DisaggregatedEvaluations}.
Once such an axis is specified, our mechanism can detect representational drift as a movement of probability mass and may help correct it through the kind of kernel-aware mitigation developed in this work.

\section{Conclusion}
\label{sec:conclusion}

In summary, we identify the problem of physical misgeneralization in sequence models, and reveal its mechanism to be one in which local deviations induced by the model are propagated through measurements of physically meaningful quantities.
We situate the phenomenon in the relevant literature, formalize the problem, propose a novel mechanistic perspective, and show that it has predictive power in accurately forecasting how real trained diffusion models transport density over physical quantities.
Our experimental methodology involves designing synthetic physical sequence tasks that give us full control over the full data generation process, spanning the prior over the physical quantity, the trajectory-generating rule, and the measurement method for recovery.
Through extensive experiments with these synthetic systems and applied vignettes like double-pendulum motion simulation and Maze2D navigation, we validate the mechanistic interpretation.
We also discuss and compare major categories of possible mitigation strategies, and showcase how the kernel may help inform useful data representation designs that reduce drift.
Taken together, our work contributes significant new understanding that we hope will be valuable for making physical systems more safe and transparent.

\begin{ack}
KN thanks Harvard University and NTT Research for jointly funding this work, Comcast for the opportunity to advance this project through an internship, Daniel Filan and the MATS Program~(Machine Alignment, Transparency, and Security; Cohort 7.0) for supporting the early stages of this research, and Sophie L. Wang for providing valuable feedback on the writing.
\end{ack}

\bibliographystyle{plainnat}
\bibliography{references}

\appendix
\clearpage

\section{Experimental Details}
\label{sec:appendix}

\newcommand{\cmark}{\ensuremath{\checkmark}}
\newcommand{\xmark}{\ensuremath{\times}}

\subsection{Setups and Hyperparameters}
\label{sec:app:readout}

\paragraph{Data normalization.}
Unless stated otherwise, we linearly map each coordinate of each trajectory into $[-1,1]$ before the diffusion model is trained, and map it back into the range of physical coordinates before we measure the quantity.
Thus, we give the generator normalized sequences, whereas we recover the quantity in the physical units used to define the dataset.

\paragraph{Model and optimizer.}
The default denoiser for the synthetic systems uses channel multipliers $(1,2,2,2)$, a sinusoidal embedding of the diffusion timestep with dimension $32$, Mish activations, group normalization, and no attention blocks.
We train with AdamW using learning rate $3\cdot 10^{-4}$, weight decay $10^{-5}$, batch size $1024$, and EMA decay $0.999$.

\paragraph{Plotting conventions.}
For each one-dimensional plot over quantity values, we use one grid over quantity values for all curves shown in the panel.
When a curve estimates density, we use the same KDE convention across all curves in that figure.
For density curves, we compute total variation on the plotted grid after normalizing each curve to integrate to $1$.

\paragraph{Compute.} We train our models using an in-house desktop PC with 6 NVIDIA A6000 GPUs with 48GB of VRAM each.
The synthetic experiments take roughly 24 GPU-hours; the applied double-pendulum, Maze2D, and text-to-speech experiments take roughly 120 GPU-hours.

\subsection{Recovering Physical Quantities}
\label{sec:app:quantity-recovery}

\subsubsection{Synthetic Setups}
\label{sec:app:synthetic-recovery}
Let $\{r_j\}_{j=1}^J$ be the grid over quantity values, and let $g(r_j)$ be the trajectory generated at $r_j$.
To recover $r$ from an observed trajectory $x$, we compare $x$ against each reference trajectory.
With
\begin{equation}
e_j(x)=\frac{1}{H}\|x-g(r_j)\|_2^2,
\end{equation}
we define
\begin{equation}
\begin{aligned}
\Rec(r_j\mid x)
&\propto
\exp\!\left(-\frac{e_j(x)}{2\sigma^2(x)}\right)\pi(r_j),\\
\sigma^2(x)
&=
\max\!\left(\min_j e_j(x),\sigma_{\min}^2\right).
\end{aligned}
\label{eq:posterior}
\end{equation}
The adaptive bandwidth's positive floor prevents a near-exact match to a reference trajectory from collapsing the posterior to a degenerate point mass.
When we report a single recovered quantity value rather than a posterior over $r$, we use the posterior mode.

\subsubsection{Double-pendulum Energy}
\label{sec:app:pendulum}
We choose initial conditions to realize the sampled energy and rescale angular velocities after each integration step to maintain that energy.

For a double-pendulum angle trajectory sampled with $\Delta t=0.01$, we recover energy by first estimating angular velocities with central differences,
\begin{equation}
\dot q_t=\frac{q_{t+1}-q_{t-1}}{2\Delta t},
\end{equation}
and compute
\begin{equation}
E_t
=
\frac{1}{2} \dot q_t^\top M(q_t)\dot q_t + V(q_t)-V_{\min},
\end{equation}
at each interior timestep, where $V_{\min}$ is the potential energy at the downward equilibrium.
We then take the median over time.
The median suppresses occasional spikes from finite differences caused by local irregularities in generated angles.

\subsubsection{Maze2D Path Length}
\label{sec:app:maze2d}
For a Maze2D position trajectory $x=(q_0,\ldots,q_{H-1})$ with $H=128$, we recover the quantity by computing total path length
\begin{equation}
\ell(x)=\sum_{t=0}^{H-2}\|q_{t+1}-q_t\|_2.
\end{equation}
Equivalently, because the horizon and timestep are fixed, in our implementation we bin by trajectory speed $\ell(x)/((H-1)\Delta t)$.

For diffusion models that generate positions directly, we draw start--goal pairs from data segments and inpaint the endpoint positions with noise matched to each diffusion timestep.
We filter generated paths by connectivity through free space and step length.

\subsection{Details of the Mitigations}
\label{sec:app:mitigations}

\subsubsection{Inverse Reweighting}
\label{sec:app:inverse}
Let $b(r)\in\{1,\ldots,B\}$ denote the bin containing quantity value $r$, let $\pi_b$ be the intended mass in bin $b$, and let $\hat\pi_b$ be the mass recovered from samples from the baseline model.
For a fixed training set $\{x_i\}_{i=1}^N$ with recovered values $r_i$, we sample trajectory $i$ in the next training run with probability proportional to
\begin{equation}
w_i=\frac{\pi_{b(r_i)}}{\max\{\hat\pi_{b(r_i)},10^{-6}\}}.
\end{equation}
We then normalize $\{w_i\}_{i=1}^N$ to sum to $1$ before drawing mini-batches.
For the synthetic systems, since data are generated from known quantity values, we implement the intervention in practice by first sampling a bin from the inverse prior
\begin{equation}
\tilde\pi_b \propto \frac{1}{\max\{\hat\pi_b,0.10/B\}},
\end{equation}
then sampling $r$ uniformly within that bin and generating the corresponding trajectory from the ground-truth rule.
We use $B=64$ for this synthetic inverse prior, $B=40$ for double-pendulum, and $B=20$ for Maze2D.

\subsubsection{Conditional Modeling}
\label{sec:app:conditional}
For each conditional model, we first map the requested quantity value $r$ to a sinusoidal embedding with the same dimension as the diffusion timestep embedding.
Writing $u=(r-r_{\min})/(r_{\max}-r_{\min})$, the embedding contains dyadic Fourier features
\begin{equation}
c(r)=
\bigl(
\sin(2\pi 2^0u),\ldots,\sin(2\pi 2^{m-1}u),
\cos(2\pi 2^0u),\ldots,\cos(2\pi 2^{m-1}u)
\bigr),
\end{equation}
truncated or padded to the denoiser embedding dimension.
We add $c(r)$ to the timestep embedding, use the same vector to modulate each residual block with FiLM~\citep{Perez2018FiLM}, and also project it to a constant sequence of condition channels concatenated to the input.
At generation time, we draw requested quantity values from the intended prior and sample with classifier-free guidance weight $1$.
For Maze2D, the start and goal endpoints are clamped via inpainting, as in the original work.

\subsubsection{Coordinate Transform}
\label{sec:app:coordinate}
Let $\{x_i\}_{i=1}^N$ be the trajectory support used for the intervention, and let $\nu_i(r)=\Rec(r\mid x_i)$ be the recovered posterior over the physical quantity for trajectory $i$.
We draw an unlabeled code support $\{y_j\}_{j=1}^N$ from a Latin-hypercube design in $[0,1]^D$, where $D$ is the flattened input dimension of the diffusion model.
The codes define only the model-facing coordinates; we use the recovered physical quantity to choose a pairing $a:\{1,\ldots,N\}\to\{1,\ldots,N\}$ that assigns physical trajectories to code points. To choose said pairing, we first apply a local kernel of the same form as the prediction kernel to code space.
For each source code $y_j$, the kernel produces perturbed codes $\widetilde y$, and the decoder assigns $\widetilde y$ to its $k_{\mathrm{dec}}$ nearest code points with weights
\begin{equation}
\alpha_\ell(\widetilde y)
\propto
\exp\!\left(
-\frac{\|\widetilde y-y_{\ell}\|_2^2-\|\widetilde y-y_{\ell_1}\|_2^2}{2\tau}
\right),
\end{equation}
where $\ell_1$ is the nearest code point and $\tau$ is the median squared distance to the farthest neighbor among the decoder neighbors.
This construction gives a sparse matrix $M$ whose entry $M_{j\ell}$ is the expected decoder weight from source code $j$ to code cell $\ell$.
If code cell $\ell$ is paired with trajectory $x_{a(\ell)}$, then the quantity mixture locally decoded around source code $j$ is
\begin{equation}
p_j(r)=\sum_{\ell=1}^N M_{j\ell}\nu_{a(\ell)}(r).
\end{equation}
We initialize the pairing so that the number of assigned trajectories in each quantity bin matches the data, then improve it with random cross-bin swaps that reduce
\begin{equation}
\frac{1}{N}\sum_{j=1}^N
\left\|p_j-\pi\right\|_2^2.
\end{equation}
After this pairing has been chosen, we train an unconditional diffusion model using the paired codes $\{y_j\}$ as the coordinate representation of the data.
At generation time, a sampled code is decoded by the same $k_{\mathrm{dec}}$-nearest-neighbor weights over paired code cells, either by averaging the recovered posteriors for evaluation or by sampling one paired support trajectory according to those weights when we need an explicit trajectory.
For Maze2D, we sample a support trajectory and inpaint coordinates at both ends of its paired code, with noise matched to the diffusion timestep.

\paragraph{Maze2D validity and diversity checks.}
Here, we carry out several tests to ensure that the typical movement, overall sample quality, and overall sample diversity are not harmed by the transform mitigation.
Under these checks, the transform maintains the relevant desiderata while reducing drift over the measured quantity: median step lengths, distances to the nearest data trajectories, and maze occupancy remain comparable to those of the data.
This addresses the potential worry that the coordinate transform is changing the generator interface in an unfair way that trades off these aforementioned desiderata.

\begin{center}
\begin{tabular}{lccc}
\toprule
Desideratum & Data & Baseline & Transform \\
\midrule
Sample consistency (median step length) & 0.034 & 0.034 & 0.034 \\
Sample quality ({distance to nearest data trajectory}) & 0.053 & 0.047 & 0.054 \\
Sample diversity (maze occupancy) & 0.728 & 0.153 & 0.792 \\
\bottomrule
\end{tabular}
\end{center}

\begin{itemize}
    \item \textbf{Median step length:} The median displacement between adjacent states over all retained trajectories. The more comparable this quantity is to the data, the better.
    \item \textbf{{Distance to nearest data trajectory}:} The median distance from each retained trajectory to the nearest trajectory in the data; for data, we use the leave-one-out nearest-neighbor distance. The more comparable this quantity is to the data, the better.
    \item \textbf{Maze occupancy:} The fraction of free maze cells visited by retained trajectories. The more comparable this quantity is to the data, the better.
\end{itemize}

\subsection{Code Repository}

We provide detailed instructions for how to replicate and verify the results.
We are also including the full code for this work as a GitHub repository:

\href{https://github.com/KentoNishi/neurips26-physical-misgeneralization}{https://github.com/KentoNishi/neurips26-physical-misgeneralization}

\clearpage

\section{Additional Data Visualizations}
\label{sec:app:visualizations}

\begin{figure}[H]
    \centering
	\maybeincludegraphics[width=\linewidth]{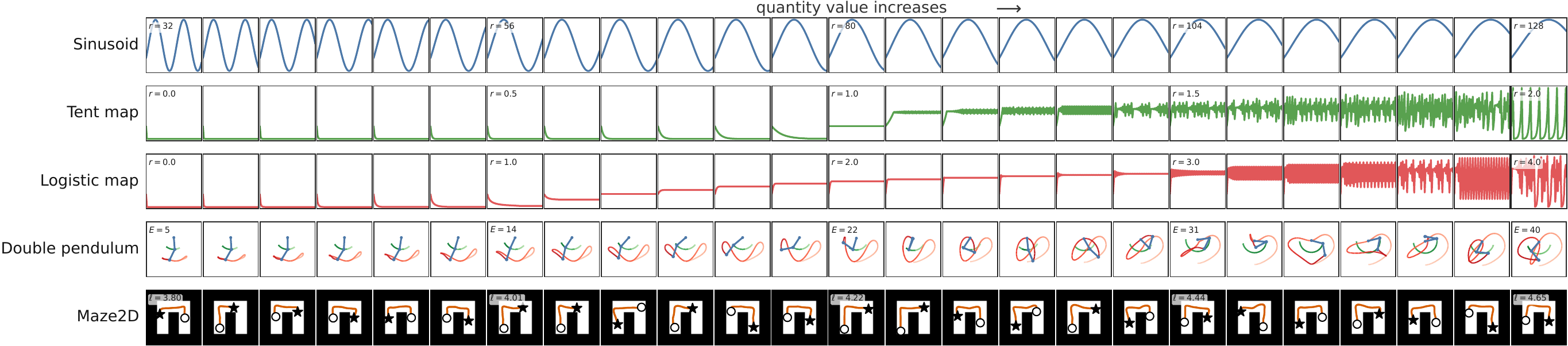}
	\caption{\textbf{Representative trajectories across the quantity ranges used to construct our datasets.}
	For each setup, we show 25 trajectories ordered from low to high quantity value.
	In the sinusoid, tent, and logistic rows, we vary the scalar quantity $r$; in the double-pendulum row, we vary total energy; in the Maze2D row, we vary path length.
	}
	\label{fig:app:synthetic-rollouts}
\end{figure}

\begin{figure}[H]
    \centering
	\maybeincludegraphics[width=\linewidth]{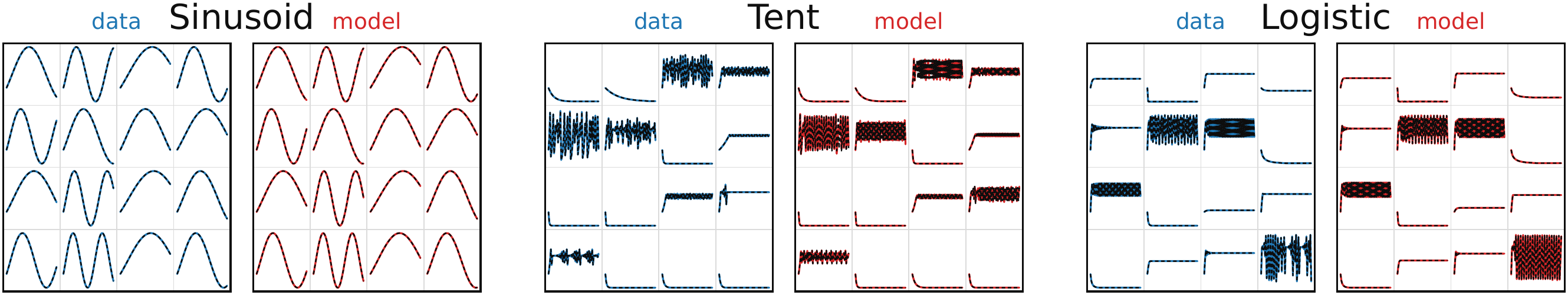}
	\caption{\textbf{Representative reconstructions for synthetic trajectories.}
	For each setup, we overlay representative trajectories with rollouts from the ground-truth data generation rule conditioned on the quantity recovered by the posterior mode.
	The colored curve is the trajectory being recovered, and the black dashed curve is the reconstructed trajectory from the posterior mode.}
	\label{fig:app:synthetic-reconstruction-overlay}
\end{figure}

\clearpage

\section{Implementation of the Data Deviation Kernel}

\subsection{Kernel for Diffusion-Based Physical Sequence Models}
\label{sec:app:kernel-impl}
The kernel must describe the local form of the diffusion sequence models' errors in trajectory space.
Since training exposes the model only to trajectories, we build this surrogate from the coordinates of those trajectories.
We emphasize that although we are hand-designing this kernel, the decisions along the way are not arbitrary. Rather, they are derived from the insights of each of the prior works we cited in relation to diffusion model phenomena. Specifically, our use of local neighborhoods and Gaussian noise follows from these prior works:
\begin{itemize}
    \item \citet{Kamb2024DiffusionCreativity} show that convolutional diffusion models recombine fragments from training data at local scales. The kernel's nearest neighbors term captures this.
    \item \citet{Everaert2024SignalLeak} shows that diffusion models are biased towards data seen in training. This justifies instantiating local recombination of training data.
    \item Finally, \citet{Aithal2024ModeInterpolation,Song2025SelectiveUnderfitting,Zhao2023DecisionStacks} each show that diffusion models that denoise from Gaussian noise can be thrown off the desired manifold during the denoising process. This justifies using perturbations beyond the empirical replacements; we use Gaussian noise to fill missing variance.
\end{itemize}

\paragraph{Defining the local recombination neighborhood.} Let $x=(x_0,\ldots,x_{H-1})$ denote a trajectory, with $x_t\in\mathbb R^d$.
We index one local piece by $u$; in our experiments, $u$ is a timestep.
Let $x_u$ denote the state at that timestep, and let $\mathcal B_u$ be the set of pieces from data at the same sequence location.
Let $d_u$ be the dimension of this local piece.
For a local scale $\sigma_u(x)>0$ set by the kernel strength $\sigma$, assign each candidate piece the Gaussian local weight
\begin{equation}
\omega_u(z\mid x)
=
\exp\left(-\frac{\|z-x_u\|_2^2}{2\sigma_u(x)^2}\right),
\qquad z\in\mathcal B_u.
\end{equation}
We then choose the empirical support of the draw using a density-stabilized Gaussian neighborhood.
The support radius should approximate a full Gaussian draw when the data resolves the local tails, while remaining conservative when only the local core is reliably sampled.
Thus, $\alpha=3$ represents the effectively full local Gaussian neighborhood, and $\alpha=1$ represents its core.
For $\alpha\in\{1,3\}$, let
\begin{equation}
\mathcal N_u^\alpha(x)
=
\left\{z\in\mathcal B_u:\|z-x_u\|_2\le \alpha\sigma_u(x)\right\}.
\end{equation}
We measure the local resolution by the number of data pieces inside $\mathcal N_u^1(x)$.
We set the cutoff for calling a neighborhood dense by taking the median of these one-scale counts for each system, then using the largest power of $2$ not exceeding the median of those medians.
When this one-scale neighborhood is densely resolved, we use the widest support $\mathcal N_u^3(x)$; when it is sparse, we use the tight support $\mathcal N_u^1(x)$.
\begin{remark}
Intuitively, this lets thin but densely sampled one-dimensional supports use a wider Gaussian neighborhood, while higher-dimensional or constrained supports stay more local.\hfill $\triangle$
\end{remark}
For those higher-dimensional or constrained supports, the radius can contain very different numbers of observed pieces across locations; so, we use the density-stabilized radius to estimate the typical number of pieces in the local Gaussian neighborhood, and implement the draw with that many nearest pieces, still weighted by the Gaussian distance weights above.
If this empirical neighborhood is empty, we use the nearest observed piece.
This gives the neighborhood $\mathcal N_u(x)$ used by the draw.

\paragraph{Drawing the replacement fragment from the neighborhood.} We then perturb the local piece by drawing a replacement $Z_u$ from this neighborhood,
\begin{equation}
\Pr(Z_u=z\mid x)
=
\frac{\omega_u(z\mid x)}{\sum_{z'\in\mathcal N_u(x)}\omega_u(z'\mid x)},
\qquad z\in\mathcal N_u(x).
\end{equation}
In the dense one-dimensional case, this draw is carried out by sampling a truncated Gaussian offset and snapping to the nearest observed piece on the discretized support.
When a task imposes additional feasibility requirements, such as fixed endpoints, local continuity, or maze free-space constraints, the same local draw is restricted or projected according to those requirements as the perturbed trajectory is assembled.
\begin{remark}
For Maze2D, these restrictions keep candidate replacements compatible with the fixed endpoints used by the inpainting setup and the surrounding path.\hfill $\triangle$
\end{remark}
For the 1D U-Net diffusion models we use, the replacement is resampled with average run length set by the first-layer convolutional kernel size, so sampled pieces persist across short local runs.

\paragraph{Incorporating continuous noise.} The draw above gives a discrete empirical perturbation.
However, the kernel should also be able to induce continuous variation around this recombination, because a diffusion error need not land exactly on another observed piece.
If the empirical draw from earlier already supplies the desired coordinate-wise variance, the residual variance we must add is zero; otherwise, we represent this residual variation with a Gaussian residual $\epsilon_u$.
Let the coordinate-wise variance introduced by empirical draws from these neighborhoods be
\begin{equation}
v_u
=
\operatorname{Var}\!\left(Z_u-x_u\right),
\end{equation}
where this variance is estimated over the sampled empirical perturbations and $v_u\in\mathbb R^{d_u}$.
When the empirical draw contributes less than the target variance $\sigma_u(x)^2$ in a coordinate, the remaining variance is added as
\begin{equation}
\epsilon_u\sim
\mathcal N\left(
0,
\operatorname{diag}\!\left(\left[\sigma_u(x)^2\mathbf 1-v_u\right]_+\right)
\right).
\end{equation}
In either case, the perturbed local piece is $\widetilde{x}_u=Z_u+\epsilon_u$.

\paragraph{The final data deviation kernel.} Given the collection of local draws, we assemble a perturbed trajectory and apply the admissibility map $\Pi_{\mathcal A}$.
This map is the identity when no projection is needed; for bounded coordinates, it clips the perturbed trajectory back to the valid coordinate range.
\begin{remark}
For Maze2D, it additionally preserves the fixed endpoints used by the inpainting setup and projects interior positions into the valid free space of the maze.
\hfill$\triangle$
\end{remark}
The final data deviation kernel is then the following conditional distribution:
\begin{equation}
D_{\mathrm{arch}}^\sigma(\delta\mid x)
\quad\text{where}\quad
x+\delta=\Pi_{\mathcal A}(\widetilde{x}).
\end{equation}
The functional form of the kernel uses neither samples generated by the model, nor residual directions, nor histograms recovered from samples from the model, nor separately trained predictors.

\clearpage

\subsection{{Model-Calibrated Comparisons Across Training}}
\label{sec:app:checkpoint-predictions}

\begin{figure}[H]
    \centering
    \includegraphics[width=\linewidth]{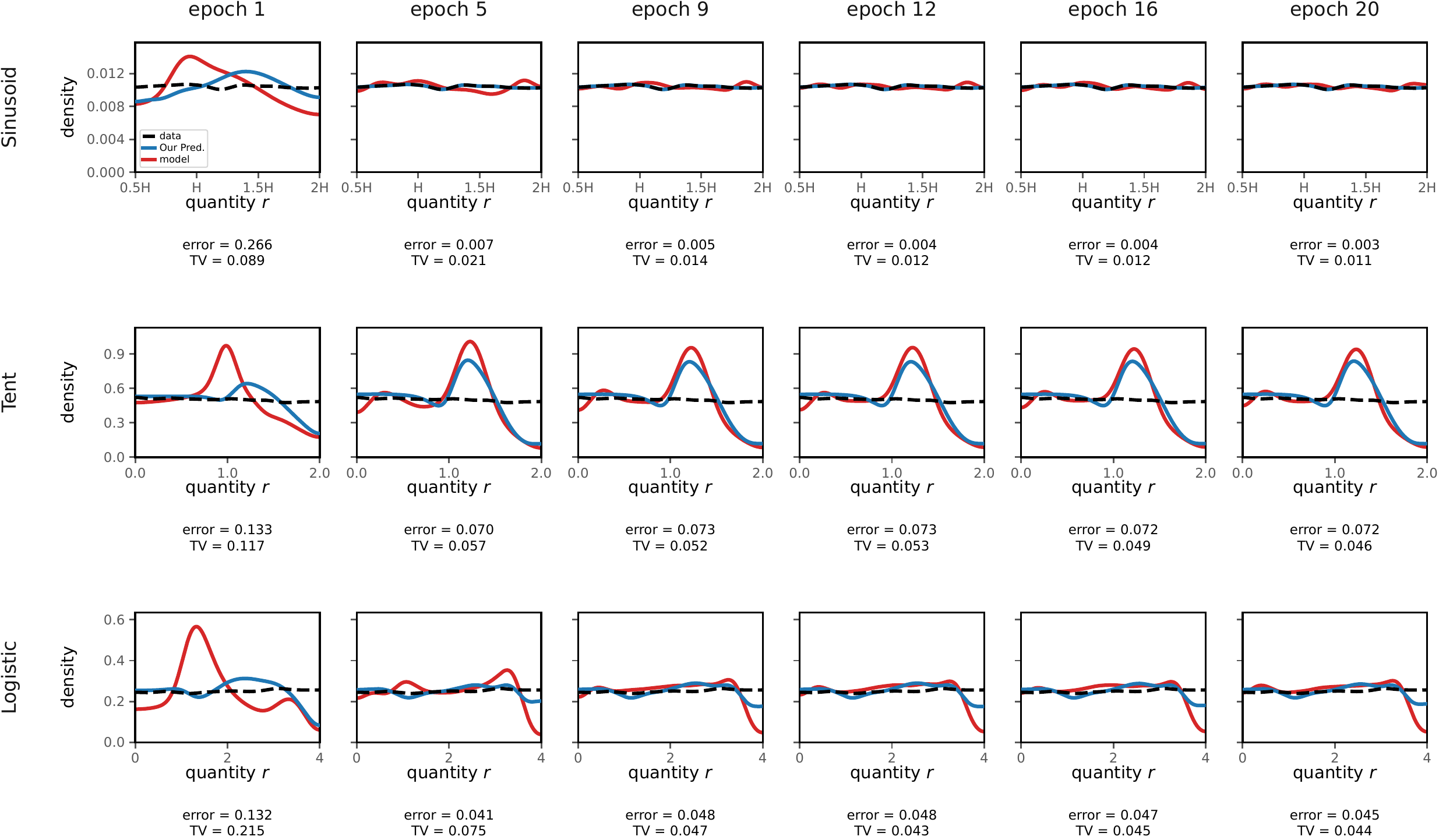}
    \caption{{\textbf{Model-calibrated predictions across training.}
    We show six evenly spaced checkpoints, pairing each model marginal with a prediction interpolated from the independent kernel sweep at that checkpoint's measured data-space error.
    We report this error and the prediction-to-model $\TV$ below each panel.
    Across all $20$ checkpoints per system, the median $\TV$ [IQR] is $0.013$ [$0.012, 0.018$] for sinusoid, $0.052$ [$0.050, 0.055$] for tent, and $0.046$ [$0.044, 0.061$] for logistic.}}
    \label{fig:app:checkpoint-progression}
\end{figure}

{In \Sref{sec:experiments}, we showed that the kernel closely predicts quantity drift for the final checkpoints of our training runs.
However, those models already fit the trajectory data well.
Here, we ask whether the same error--drift relationship continues to predict quantity drift across a much wider range of model fitness.
To test this, we evaluate all $20$ saved checkpoints per synthetic system; each supplies only its aggregate data-space error while the kernel independently supplies the perturbation structure~(\Fref{fig:app:checkpoint-progression}). Predictions are accurate from undertraining through convergence.
}

\clearpage

\subsection{Scale of the Deviations}
\label{sec:app:scale}
{In \Aref{sec:app:checkpoint-predictions}, we matched the kernel's induced data-space error to error of each checkpoint.
We next ask how sensitive the predicted quantity distribution is to the choice of deviation scale $\sigma$.
}

For the calibrated synthetic predictions, we first estimate how the model's error magnitude varies over the recovered quantity.
The local scale follows this profile, with its overall magnitude calibrated to the aggregate reconstruction error of model samples.
Here, we vary $\sigma$ while holding profiles fixed.

\begin{figure}[H]
    \centering
	\begin{subfigure}[t]{1\linewidth}
		\vspace{0pt}
		\maybeincludegraphics[width=\linewidth]{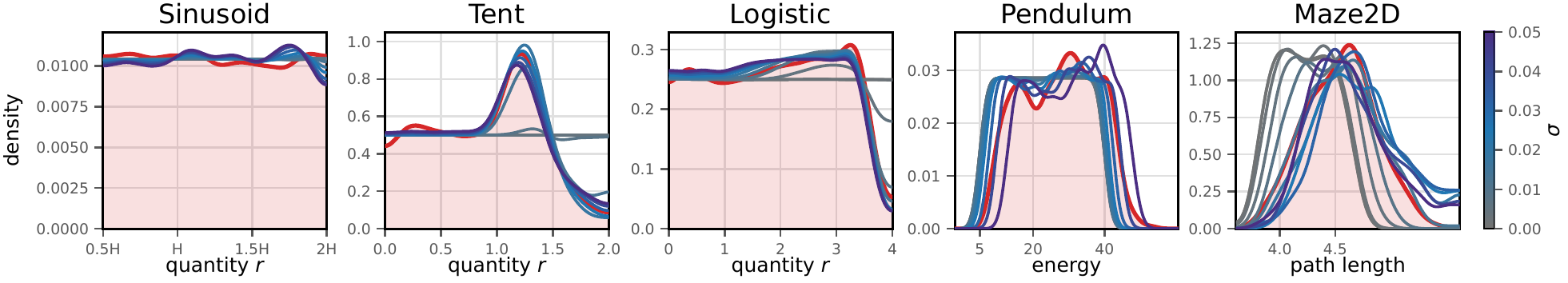}
	\end{subfigure}
	\begin{subfigure}[t]{1\linewidth}
    \centering
		\vspace{4pt}
		\begingroup
\setlength{\tabcolsep}{1.55pt}
\renewcommand{\arraystretch}{0.86}
\centering
\begin{tabular}{@{}rccccc@{}}
\toprule
$\sigma$ & Sin. & Tent & Log. & Pend. & M2D \\
\midrule
\textcolor[HTML]{737373}{0} & \textcolor[HTML]{737373}{0.012} & \textcolor[HTML]{737373}{0.159} & \textcolor[HTML]{737373}{0.066} & \textcolor[HTML]{737373}{0.163} & \textcolor[HTML]{737373}{0.378} \\
\textcolor[HTML]{717374}{.0005} & \textcolor[HTML]{717374}{0.012} & \textcolor[HTML]{717374}{0.159} & \textcolor[HTML]{717374}{0.066} & \textcolor[HTML]{717374}{0.163} & \textcolor[HTML]{717374}{0.376} \\
\textcolor[HTML]{6C7378}{.002} & \textcolor[HTML]{6C7378}{0.012} & \textcolor[HTML]{6C7378}{0.159} & \textcolor[HTML]{6C7378}{0.066} & \textcolor[HTML]{6C7378}{0.163} & \textcolor[HTML]{6C7378}{0.356} \\
\textcolor[HTML]{64747F}{.0045} & \textcolor[HTML]{64747F}{0.012} & \textcolor[HTML]{64747F}{0.159} & \textcolor[HTML]{64747F}{0.042} & \textcolor[HTML]{64747F}{0.163} & \textcolor[HTML]{64747F}{0.245} \\
\textcolor[HTML]{587487}{.008} & \textcolor[HTML]{587487}{0.012} & \textcolor[HTML]{587487}{0.151} & \textcolor[HTML]{587487}{0.015} & \textcolor[HTML]{587487}{0.160} & \textcolor[HTML]{587487}{0.146} \\
\textcolor[HTML]{497594}{.0125} & \textcolor[HTML]{497594}{0.012} & \textcolor[HTML]{497594}{0.042} & \textcolor[HTML]{497594}{0.017} & \textcolor[HTML]{497594}{0.155} & \textcolor[HTML]{497594}{0.031} \\
\textcolor[HTML]{3676A2}{.018} & \textcolor[HTML]{3676A2}{0.013} & \textcolor[HTML]{3676A2}{0.030} & \textcolor[HTML]{3676A2}{0.022} & \textcolor[HTML]{3676A2}{0.139} & \textcolor[HTML]{3676A2}{0.127} \\
\textcolor[HTML]{2177B3}{.0245} & \textcolor[HTML]{2177B3}{0.015} & \textcolor[HTML]{2177B3}{0.020} & \textcolor[HTML]{2177B3}{0.025} & \textcolor[HTML]{2177B3}{0.113} & \textcolor[HTML]{2177B3}{0.176} \\
\textcolor[HTML]{2B63A6}{.032} & \textcolor[HTML]{2B63A6}{0.019} & \textcolor[HTML]{2B63A6}{0.019} & \textcolor[HTML]{2B63A6}{0.027} & \textcolor[HTML]{2B63A6}{0.077} & \textcolor[HTML]{2B63A6}{0.177} \\
\textcolor[HTML]{3A4995}{.0405} & \textcolor[HTML]{3A4995}{0.023} & \textcolor[HTML]{3A4995}{0.030} & \textcolor[HTML]{3A4995}{0.030} & \textcolor[HTML]{3A4995}{0.051} & \textcolor[HTML]{3A4995}{0.123} \\
\textcolor[HTML]{4B2E83}{.05} & \textcolor[HTML]{4B2E83}{0.027} & \textcolor[HTML]{4B2E83}{0.038} & \textcolor[HTML]{4B2E83}{0.033} & \textcolor[HTML]{4B2E83}{0.133} & \textcolor[HTML]{4B2E83}{0.097} \\
\bottomrule
\end{tabular}
\endgroup

	\end{subfigure}
	\caption{\textbf{The deviation scale controls how strongly local trajectory errors are expressed.}
	For each system, we sweep the kernel's absolute scale $\sigma$ from zero upward.
	Here, the solid red line is the trained model.
    We see that increasing the scale amplifies the redistribution of probability.
    Notably, the predicted curves are very stable, and the flat baseline gradually morphs into the shape of the actual models' distributions as $\sigma$ increases.
    Then, as expected, the prediction is over-amplified and overshoots the target model when $\sigma$ is set too large.
    We report the exact TV values of the predicted distribution compared to the corresponding trained model in the table immediately below.
    }
	\label{fig:app:kernel-scale-sweep}
\end{figure}

\clearpage

\subsection{{Model-Free Scale Selection}}
\label{sec:app:kernel-only}

{In \Aref{sec:app:scale}, we observed stable predicted shapes over a range of scales, with the synthetic models' error profiles held fixed.
We now ask whether we can also select a single prediction without information from the target model.
This is less straightforward because models at different levels of fit need not share one useful predicted quantity distribution.
Fortunately, we can select a scale by exploiting the observed stability of kernel-based predictions.
}

\begin{figure}[H]
    \centering
    \includegraphics[width=\linewidth]{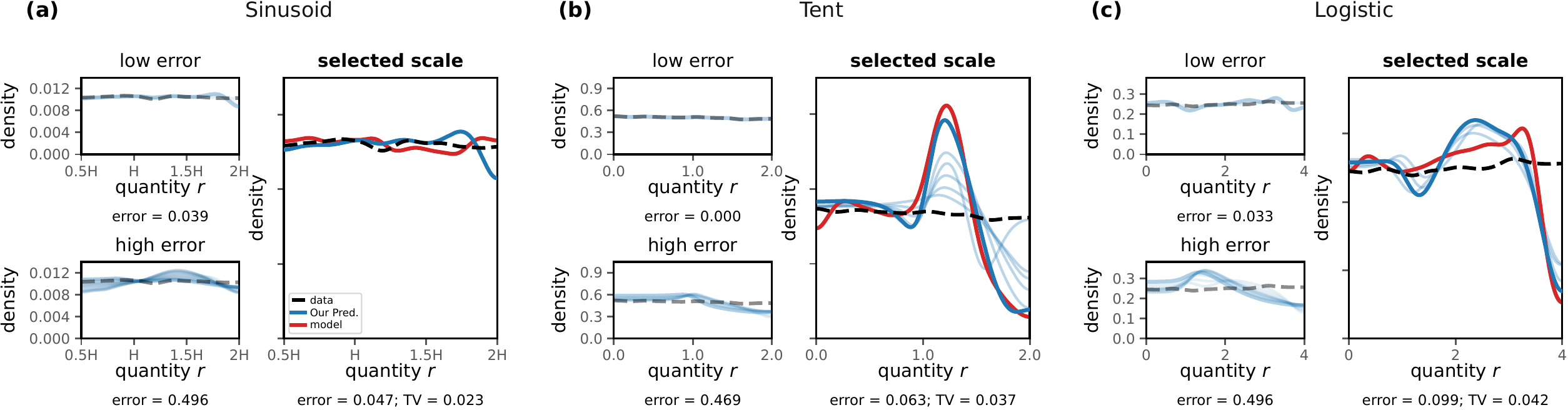}
    \caption{{\textbf{Predictions across the model-free sweep.}
    We show all sweep points, grouped into predictions before drift onset, from onset through the selected basin, and beyond that basin.
    On the right of each group, we highlight the prediction interpolated at the selected scale and overlay the final model marginal; fainter blue curves show the other sweep points in each regime.
    Below each plot, we report the induced data-space error of the highlighted prediction, averaged across independent batches; for the selected prediction, we also report its $\TV$ to the model.}}
    \label{fig:app:model-free-progression}
\end{figure}

{To do this, we vary a common scale across trajectories using only the empirical replacement component, ordering predictions by induced data-space error.
As error grows, the sweep passes through three phases: (1) a data-like regime before drift, (2) a stable nontrivial drift shape, and (3) saturation under excessive corruption.
We select the first stable basin in phase 2, identified by the first local minimum of a joint score for adjacent changes in normalized signed drift shape and the full predicted marginal~(\Fref{fig:app:model-free-progression}).
}

\begin{table}[H]
\centering
\begin{tabular}{lrrrr}
\toprule
{System} & {Kernel-only error [95\% CI]} & {Target error} & {$\TV$, kernel-only} & {$\TV$, calibrated} \\
\midrule
{Sinusoid} & {0.0468 [0.0349, 0.0469]} & {0.0032} & {0.023} & {0.012} \\
{Tent} & {0.0630 [0.0628, 0.0633]} & {0.0717} & {0.037} & {0.019} \\
{Logistic} & {0.0995 [0.0660, 0.0997]} & {0.0461} & {0.042} & {0.017} \\
\bottomrule
\end{tabular}

\vspace{0.5em}
\caption{{\textbf{Kernel-only selection and model-calibrated comparison.}
Errors are measured in data space; $\TV$ compares predictions with the final model marginals in \Fref{fig:synthetic-kernel-pred}.
We report $95\%$ percentile intervals from $2{,}000$ bootstrap resamples of independent batches, repeating the complete selection procedure for each resample.}}
\label{tab:app:kernel-only}
\vspace{-2em}
\end{table}

\clearpage

\subsection{{Gaussian Ablation and Validation Against Model Errors}}
\label{sec:app:kernel-validation}

{Having examined how predictions change with model fitness and deviation scale, we next ask what diffusion-specific structure contributes beyond generic noise.
As a sanity check, we compare predictions from additive Gaussian noise alone with those from the full kernel at matched data-space error, using the same recovery and model marginals as in \Fref{fig:synthetic-kernel-pred}.
}

\begin{table}[H]
\centering
\begin{tabular}{lrr}
\toprule
{System }& {Gaussian $\TV$ }& {Full kernel $\TV$ }\\
\midrule
{Sinusoid }& {0.014 }& {0.013 }\\
{Tent }& {0.046 }& {0.020 }\\
{Logistic }& {0.050 }& {0.016 }\\
\bottomrule
\end{tabular}
\vspace{0.5em}
\caption{{\textbf{{Distances between predicted and model marginals} at matched data-space error.}}}
\vspace{-1.5em}
\label{tab:app:gaussian-ablation}
\end{table}

{For sinusoid, the two predictions have nearly equal $\TV$; for tent and logistic, we obtain lower $\TV$ with the full kernel.
}

{{To check whether actual model errors point toward local fragments from the data, we also perform a \emph{post hoc validation} using residuals between denoised and original data trajectories.}
Within length-$3$ windows matching the kernel's replacement runs, we measure the \emph{maximum cosine similarity} of each residual to directions toward nearby data fragments at the same locations.
We compare model, kernel, and isotropic Gaussian deviations using the same candidate fragments and matching their aggregate root-mean-square magnitudes.
}

\begin{table}[H]
\centering
\begin{tabular}{lrrr}
\toprule
{System }& {Actual }& {Kernel }& {Gaussian }\\
\midrule
{Sinusoid }& {0.962 }& {0.972 }& {0.764 }\\
{Tent }& {0.683 }& {0.771 }& {0.577 }\\
{Logistic }& {0.258 }& {0.458 }& {0.145 }\\
\bottomrule
\end{tabular}
\vspace{0.5em}
\caption{{\textbf{Alignment with directions toward nearby data fragments.}
{Entries are median maximum cosine similarities.}}}
\label{tab:app:fragment-alignment}
\vspace{-1.5em}
\end{table}

{The results above collectively show that denoising errors point toward {local fragments from the data} more than isotropic Gaussian noise does.
}

\clearpage

\section{Sensitivity of the Synthetic Families}
\label{sec:app:lyapunov-proofs}

The synthetic families let us compare a learnable baseline against two systems where local trajectory errors can grow along the rollout. Here, we showcase and formally prove this claim.

\subsection{State Geometry and Lyapunov Exponents}

\begin{figure}[H]
    \centering
	\maybeincludegraphics[width=\linewidth]{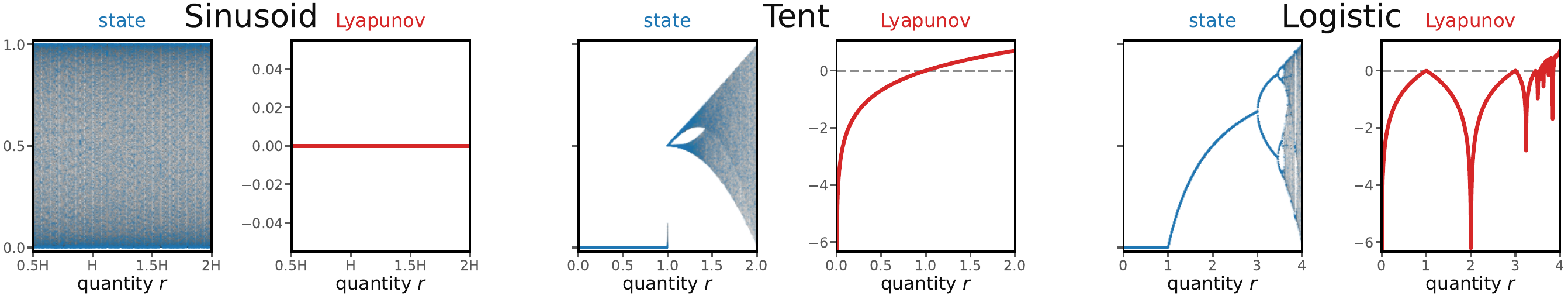}
	\caption{\textbf{The synthetic families differ in how local trajectory errors grow along the rollout.}
	Within each system, the left panel shows the states reached across the range of quantity values, and the right panel shows the corresponding Lyapunov exponent.
	The sinusoid has no expanding recurrence, whereas the tent and logistic maps include regimes where nearby rollouts separate rapidly; We examine how this state sensitivity relates to prediction residuals in~\Aref{sec:app:lyapunov-residual}.}
	\label{fig:tent}
\end{figure}

\subsection{Lyapunov Proofs for the Synthetic Families}

\subsubsection{Sinusoidal Family}
\label{sec:app:lyapunov-sinusoid}
For the sinusoidal family, a perturbation $\eta_t$ to phase has identity tangent dynamics, {since phase advances by the constant $2\pi/r$ each timestep,
}$
\eta_{t+1}=\eta_t.
$
Hence,
\begin{equation}
\lambda_H^{\mathrm{sinusoid}}
=
\frac{1}{H-1}\sum_{t=0}^{H-2}\log 1
=0.
\end{equation}
The sinusoid is a zero-growth baseline against which the tent and logistic maps can be compared.

{Also, at the fixed phase used to generate data, the mean value theorem gives $|x_t(r')-x_t(r)|\le (\pi t/32^2)|r'-r|$ for periods $r,r'\in[32,128]$.
Thus, nearby periods also produce nearby trajectories over the fixed horizon.
}

\subsubsection{Tent Map}
\label{sec:app:lyapunov-tent}
For a one-dimensional dynamical system $x_{t+1}=f_r(x_t)$, a perturbation $\eta_t$ evolves according to
\begin{equation}
\eta_{t+1}=f_r'(x_t)\eta_t,
\end{equation}
so
\begin{equation}
|\eta_{H-1}|=|\eta_0|\prod_{t=0}^{H-2}|f_r'(x_t)|.
\end{equation}
In the tent map, $f_r(x)=r\min\{x,1-x\}$ has derivative $r$ on the left branch and $-r$ on the right branch.
Thus, for trajectories that do not land exactly on the fold,
\begin{equation}
\lambda_H^{\mathrm{tent}}(r)
=
\frac{1}{H-1}\sum_{t=0}^{H-2}\log |f_r'(x_t)|
=
\log r.
\end{equation}
As $r$ increases above $1$, local perturbations are amplified exponentially along the rollout.

\subsubsection{Logistic Map}
\label{sec:app:lyapunov-logistic}
The logistic map follows the same perturbation calculation, but its derivative depends on the state reached along the rollout.
Since $f_r(x)=rx(1-x)$, we have $f_r'(x)=r(1-2x)$, and therefore
\begin{equation}
\lambda_H^{\mathrm{logistic}}(r;x_0)
=
\frac{1}{H-1}\sum_{t=0}^{H-2}\log |r(1-2x_t)|.
\end{equation}
Unlike the tent map, the orbit itself enters the product, so the exponent need not be monotone in $r$ over the full interval.
In the stable regime with a fixed point, $1<r<3$, the attracting fixed point is $x^\star=1-1/r$, giving
\begin{equation}
\lambda_\infty^{\mathrm{logistic}}(r)
=
\log |f_r'(x^\star)|
=
\log |2-r| < 0.
\end{equation}

\subsection{{Prediction Residuals and Lyapunov Exponents}}
\label{sec:app:lyapunov-residual}

{We find no straightforward relationship between the Lyapunov profiles in \Fref{fig:tent} and the recovered quantity distributions in \Fref{fig:synthetic-kernel-pred}, including when we restrict the tent map to ranges with low or high Lyapunov exponents~(\Aref{sec:app:hypothesis}).
Nevertheless, the residual between prediction and model could concentrate at quantity values with higher Lyapunov exponents.
}

{To test this, we compute the pointwise absolute density residual between calibrated prediction and model at every quantity value, rank values by local Lyapunov exponent, divide them into four equal-count groups, and measure each group's mean residual (\textit{i.e.}, Q1 has the lowest exponents; Q4, the highest).
We also report the Spearman correlation between residual and exponent.
}

\begin{table}[H]
\centering
\begin{tabular}{lrrrrr}
\toprule
{System }& {Q1 }& {Q2 }& {Q3 }& {Q4 }& {Spearman $\rho$ }[{95\% CI}] \\
\midrule
{Tent }& {0.0326 }& {0.0142 }& {0.0189 }& {0.0085 }& {-0.487 }[{-0.667, -0.129}] \\
{Logistic }& {0.0081 }& {0.0091 }& {0.0067 }& {0.0109 }& {0.141 }[{-0.419, 0.575}] \\
\bottomrule
\end{tabular}
\vspace{0.5em}
\caption{{\textbf{Prediction residuals grouped by Lyapunov exponent.}
Entries in Q1--Q4 are mean absolute density residuals.
Intervals are $95\%$ percentile intervals from $2{,}000$ bootstrap resamples of recovery posteriors for individual trajectories.}}
\label{tab:app:lyapunov-residual}
\end{table}

{Tent residuals are non-monotonic, with Q4 below Q1 by $0.0241$ and a negative rank correlation.
Logistic residuals are also non-monotonic, with Q4 above Q1 by $0.0029$ and a weak positive correlation whose interval spans zero.
Thus, residuals are not systematically concentrated in {regions with high Lyapunov exponents} across both systems.
}

\clearpage

\section{Variation Across Training Seeds}
\label{sec:app:statistics}

To check that the prediction and mitigation results are not the byproduct of one favorable training run, we repeat the central experiments with three independent training seeds.
For each run, we recover the distribution over the physical quantity and report $\TV$ distances using the same evaluation code used for the corresponding figures.
\Tref{tab:app:prediction-stability} shows that, across the synthetic suite and the two applied physical systems, the {distances between predicted and model marginals} remain small relative to the scale of the model drift.
\Tref{tab:app:mitigation-stability} also shows that $\TV$ from the intended prior is stable for both the baseline model and the mitigation strategies.
Across seeds, all our conclusions remain intact.

\begin{table}[H]
    \centering
    \small
\setlength{\tabcolsep}{3pt}
\resizebox{\linewidth}{!}{%
\begin{tabular}{lccccc}
\toprule
Metric & Sinusoid & Tent & Logistic & Pendulum & Maze2D \\
\midrule
$\TV$(data, prior) & $0.004, 0.004, 0.003$ & $0.004, 0.004, 0.003$ & $0.004, 0.004, 0.003$ & $0.040, 0.040, 0.040$ & $0.041, 0.039, 0.039$ \\
$\TV$(model, prior) & $0.026, 0.041, 0.044$ & $0.186, 0.193, 0.204$ & $0.097, 0.155, 0.133$ & $0.164, 0.153, 0.136$ & $0.361, 0.431, 0.392$ \\
$\TV$(prediction, prior) & $0.003, 0.005, 0.006$ & $0.166, 0.164, 0.170$ & $0.088, 0.083, 0.087$ & $0.164, 0.119, 0.117$ & $0.337, 0.292, 0.354$ \\
$\TV$(prediction, model) & $0.027, 0.040, 0.044$ & $0.036, 0.084, 0.056$ & $0.032, 0.108, 0.102$ & $0.056, 0.057, 0.046$ & $0.113, 0.179, 0.139$ \\
\bottomrule
\end{tabular}
}%

    \vspace{0.5em}
    \caption{\textbf{Prediction stability across training seeds.}
    Each cell lists the values from three independent training seeds.
    The $\TV$(prediction, model) row directly compares the prediction to the distribution recovered from the trained model.}
    \label{tab:app:prediction-stability}
\end{table}

\begin{table}[H]
    \centering
    \small
\setlength{\tabcolsep}{3pt}
\resizebox{\linewidth}{!}{%
\begin{tabular}{lccccc}
\toprule
Method & Sinusoid & Tent & Logistic & Pendulum & Maze2D \\
\midrule
Baseline & $0.026, 0.041, 0.044$ & $0.186, 0.193, 0.204$ & $0.097, 0.155, 0.133$ & $0.190, 0.170, 0.151$ & $0.182, 0.164, 0.258$ \\
Reweight & $0.034, 0.028, 0.058$ & $0.217, 0.196, 0.222$ & $0.202, 0.157, 0.159$ & $0.672, 0.640, 0.422$ & $0.115, 0.182, 0.199$ \\
Cond. & $0.007, 0.008, 0.008$ & $0.026, 0.016, 0.017$ & $0.012, 0.013, 0.014$ & $0.200, 0.174, 0.199$ & $0.151, 0.161, 0.136$ \\
Transform & $0.026, 0.020, 0.034$ & $0.026, 0.021, 0.036$ & $0.029, 0.021, 0.043$ & $0.029, 0.026, 0.027$ & $0.026, 0.023, 0.022$ \\
\bottomrule
\end{tabular}
}%

    \vspace{0.5em}
    \caption{\textbf{Mitigation stability across training seeds.}
    Values are $\TV$ distances from the intended prior across three independent training seeds.}
    \label{tab:app:mitigation-stability}
\end{table}

Assuming that paired differences across seeds are approximately normally distributed, we also compute one-sided paired $t$-tests.
For prediction, we define
\begin{equation}
d_s^{\mathrm{pred}}
=
\TV\!\left(\bar{\pi}_{\mathrm{model}}^{(s)},\pi\right)
-
\TV\!\left(\bar{\pi}_{\mathrm{dev}}^{(s)},\bar{\pi}_{\mathrm{model}}^{(s)}\right),
\end{equation}
where $s$ indexes the training seed.
For mitigation, we define
\begin{equation}
d_s^{\mathrm{mit}}
=
\TV\!\left(\bar{\pi}_{\mathrm{base}}^{(s)},\pi\right)
-
\TV\!\left(\bar{\pi}_{\mathrm{transport}}^{(s)},\pi\right).
\end{equation}
In both cases, the one-sided null is $\mathbb{E}[d_s]\le 0$.
Writing $\bar d$ for the mean of the three paired differences and $s_d$ for their sample standard deviation, we use
\begin{equation}
t=\frac{\bar d}{s_d/\sqrt{3}},
\qquad
p=\Pr(T_2\ge t).
\end{equation}

\begin{table}[H]
    \centering
    \small
\setlength{\tabcolsep}{6pt}
\begin{tabular}{lcc}
\toprule
System & Prediction $p$ & Transform mitigation $p$ \\
\midrule
Sinusoid & $0.378$ & $0.118$ \\
Tent & $0.005$ & $<10^{-3}$ \\
Logistic & $0.020$ & $0.019$ \\
Double-pendulum & $0.002$ & $0.003$ \\
Maze2D & $<10^{-3}$ & $0.013$ \\
\bottomrule
\end{tabular}

    \vspace{0.5em}
    \caption{\textbf{One-sided paired tests across training seeds.}
    \emph{Prediction} tests whether the {distance between predicted and model marginals} is smaller than the model's drift from the intended prior.
    \emph{Transform mitigation} tests whether the coordinate-transform intervention has lower drift than the baseline model.}
    \label{tab:app:seed-ttests}
\end{table}

These tests show that the accuracy of the prediction, as well as the effectiveness of the mitigation, is stable across seeds.
The sinusoid is the expected exception because the baseline model barely drifts; for the other systems, both tests reject the one-sided null at $p<0.05$.

\clearpage

\section{Alternative Explanations and Model Architectures}
\label{sec:app:hypothesis}

At first glance, one might suspect that physical misgeneralization is a consequence of fairly ordinary causes that are much simpler than the mechanism we proposed in the main text.
For example, the training set could be poorly sampled; the model could be too small or trained too briefly; the sampler may be poorly chosen; or, the effect could be specific to diffusion models.
For each possibility including but not limited to these aforementioned suspicions, we run an ablation with the tent map, swapping out the relevant part of recovery, data generation, training, sampling, or model choice.
Overall, we find that these alternative theories do not account for physical misgeneralization, and thus the perspective put forth by this paper merits broader attention from the community.

\begin{center}
\scriptsize
\renewcommand{\arraystretch}{1.08}
\setlength{\tabcolsep}{4pt}
\begin{tabularx}{\linewidth}{@{}l>{\raggedright\arraybackslash}Xcc@{}}
\toprule
Hyp. & Experiment & $\TV$ data & $\TV$ model \\
\midrule
H0 & Baseline diffusion model trained on trajectories from the tent map & 0.004 & 0.186 \\
H1a & Recover $r$ using the mean of the posterior & 0.018 & 0.208 \\
H1b & Recover $r$ using a discrepancy based on $\ell_1$ distance & 0.018 & 0.183 \\
H1c & Recover $r$ using both changes & 0.018 & 0.183 \\
H2a & Draw values of the quantity uniformly within bins & 0.004 & 0.191 \\
H3a & Train for fewer epochs & 0.018 & 0.195 \\
H3b & Train for more epochs & 0.018 & 0.178 \\
H3c & Repeat the baseline across seeds & 0.016 & $0.185 \pm 0.003$ \\
H4a & Widen the denoiser & 0.018 & 0.190 \\
H5a & Enlarge the convolutional kernel & 0.018 & 0.194 \\
H5b & Replace the U-Net with an MLP denoiser & 0.018 & 0.449 \\
H6a & {Replace $256$ DDPM steps with $50$ deterministic }DDIM steps & 0.018 & 0.204 \\
H7a & Train the denoiser to predict {the clean trajectory }& 0.018 & 0.361 \\
H8a & Use shorter trajectories & 0.018 & 0.170 \\
H8b & Use longer trajectories & 0.018 & 0.197 \\
H9a & Train an autoregressive Transformer & 0.018 & 0.368 \\
H9b & Train a VAE & 0.018 & 0.950 \\
H10a & Use a lower interval for $r$ & 0.015 & 0.078 \\
H10b & Use a higher interval for $r$ & 0.018 & 0.227 \\
\bottomrule
\end{tabularx}
\vspace{0.5em}
\captionof{table}{\textbf{Ablation ladder for the tent map.}
For each row, we change one proposed source of the mismatch and report the resulting distance between the distribution recovered over quantity values and the intended prior.
Substantial drift remains across variants H1--H9; H10 shows that the geometry of the data meaningfully controls the prevalence of drift, which is consistent with our mechanistic explanation wherein local deviations described by the data deviation kernel become drift after passing through the geometry of the measured quantity.}
\label{tab:app:hypothesis-v2}
\end{center}

\begin{figure}[H]
    \centering
    \newcommand{\hypminiVtwo}[2]{%
    \begingroup
    \setbox0=\hbox{\maybeincludegraphics[width=\linewidth,height=0.686\linewidth]{hypothesis_v2/#2}}%
    \leavevmode\hbox{\copy0\kern-\wd0\rlap{\raisebox{0.35em}[0pt][0pt]{\hspace{0.55em}\colorbox{white}{\scriptsize\bfseries #1}}}\kern\wd0}%
    \endgroup
}
\begin{center}
\setlength{\tabcolsep}{1pt}
\renewcommand{\arraystretch}{1.0}
\begin{tabular}{ccccc}
\begin{minipage}[t]{0.19\linewidth}\centering\hypminiVtwo{H1a}{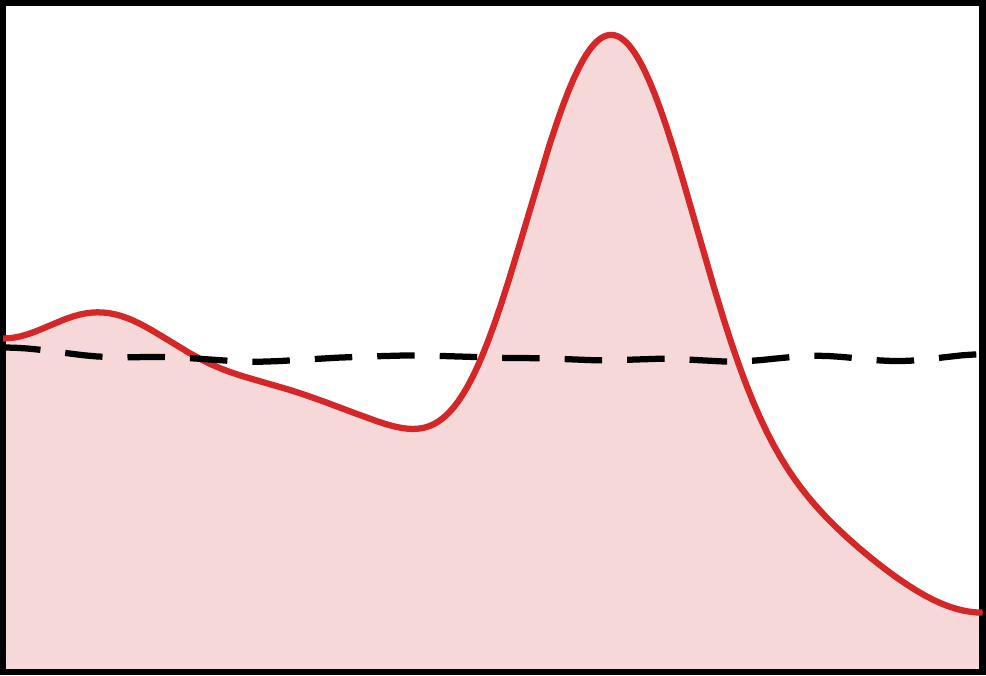}\end{minipage} & \begin{minipage}[t]{0.19\linewidth}\centering\hypminiVtwo{H1b}{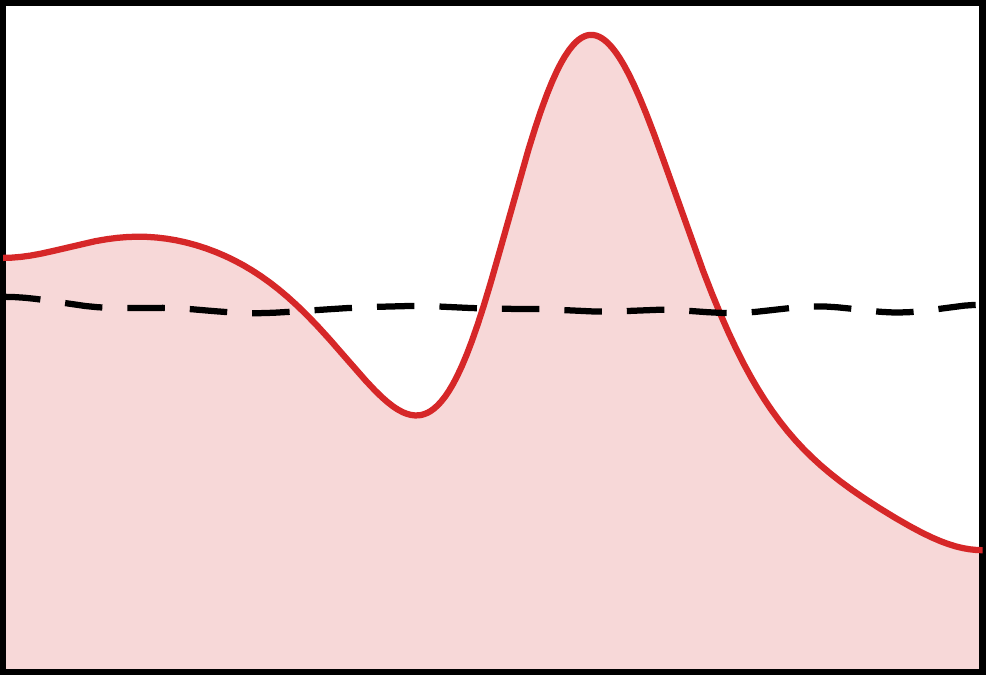}\end{minipage} & \begin{minipage}[t]{0.19\linewidth}\centering\hypminiVtwo{H1c}{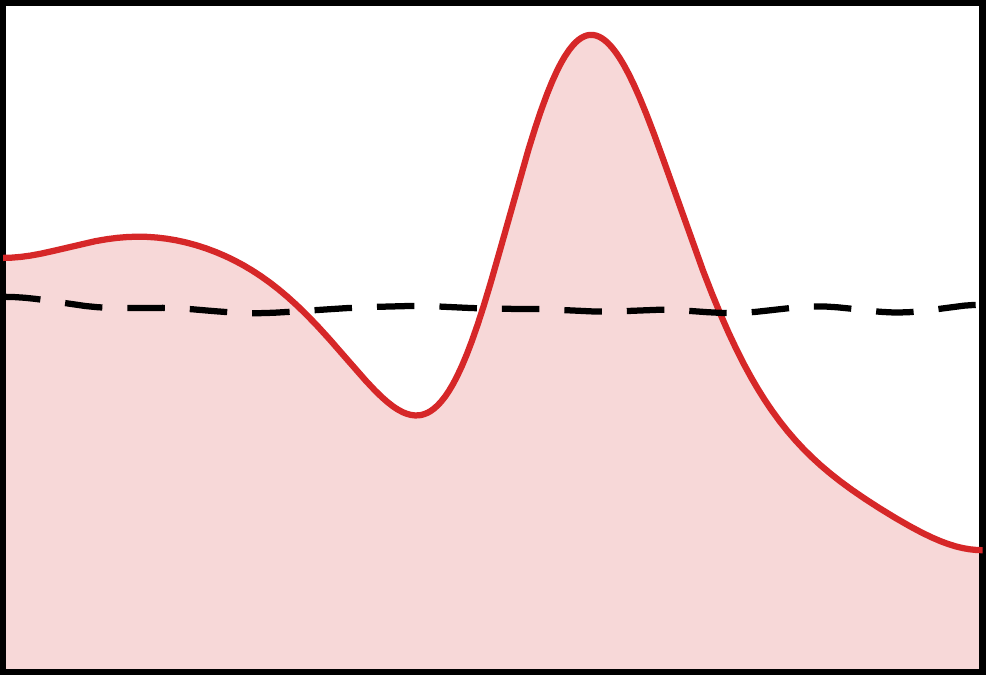}\end{minipage} & \begin{minipage}[t]{0.19\linewidth}\centering\hypminiVtwo{H2a}{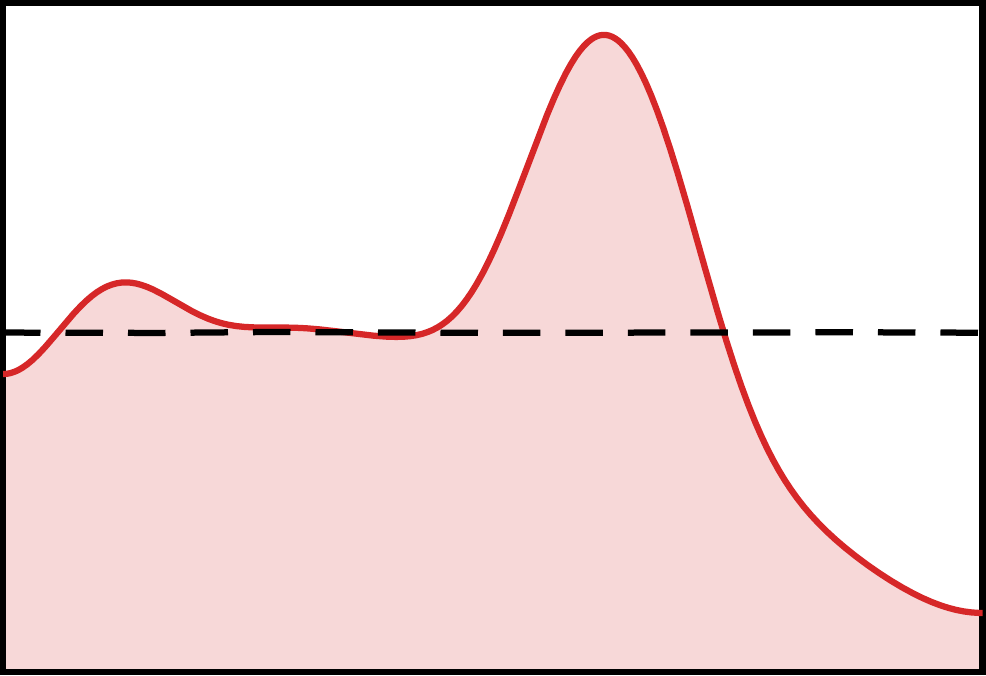}\end{minipage} & \begin{minipage}[t]{0.19\linewidth}\centering\hypminiVtwo{H3a}{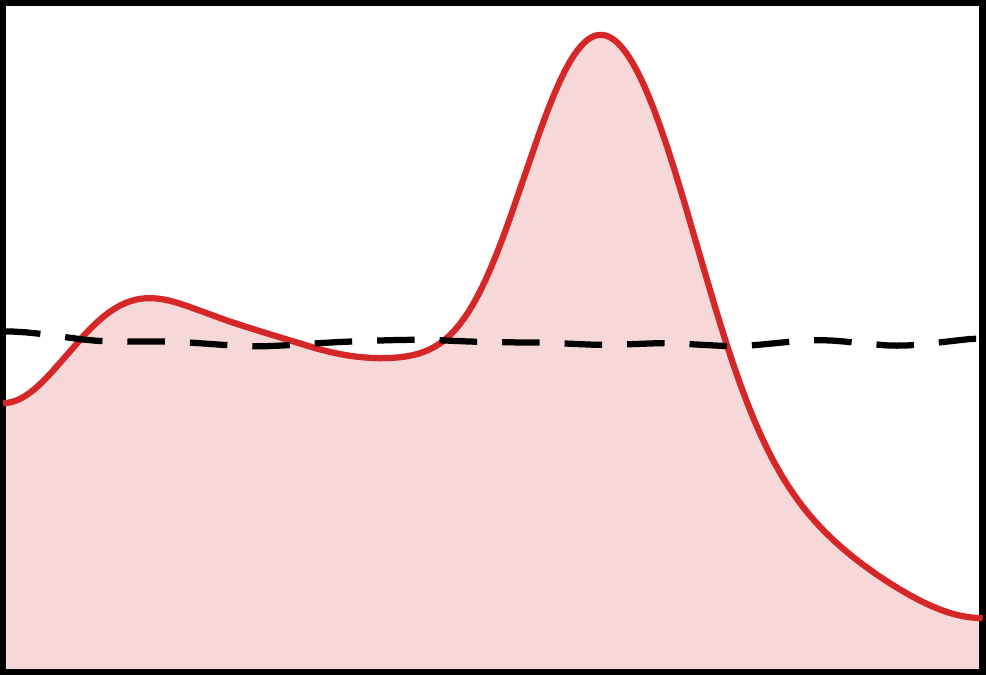}\end{minipage}\\[0.25em]
\begin{minipage}[t]{0.19\linewidth}\centering\hypminiVtwo{H3b}{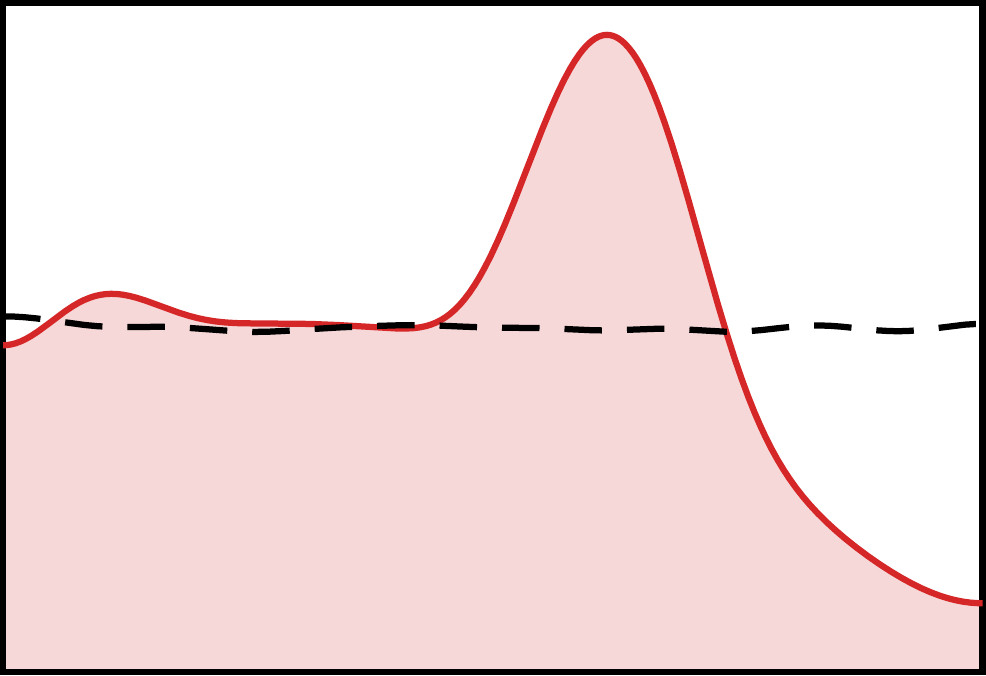}\end{minipage} & \begin{minipage}[t]{0.19\linewidth}\centering\hypminiVtwo{H3c-1}{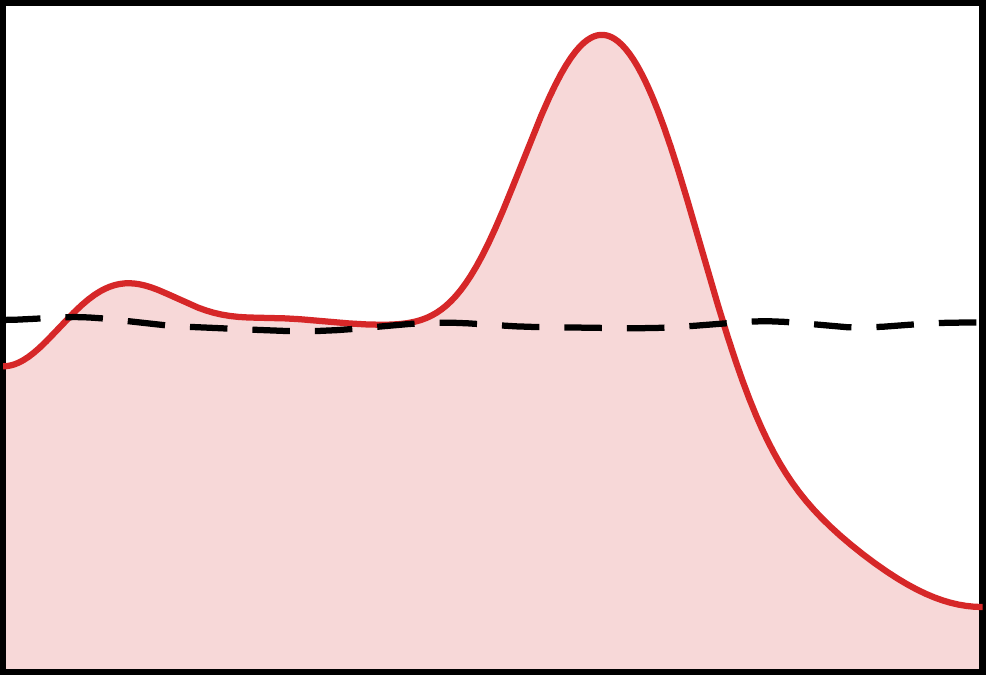}\end{minipage} & \begin{minipage}[t]{0.19\linewidth}\centering\hypminiVtwo{H3c-2}{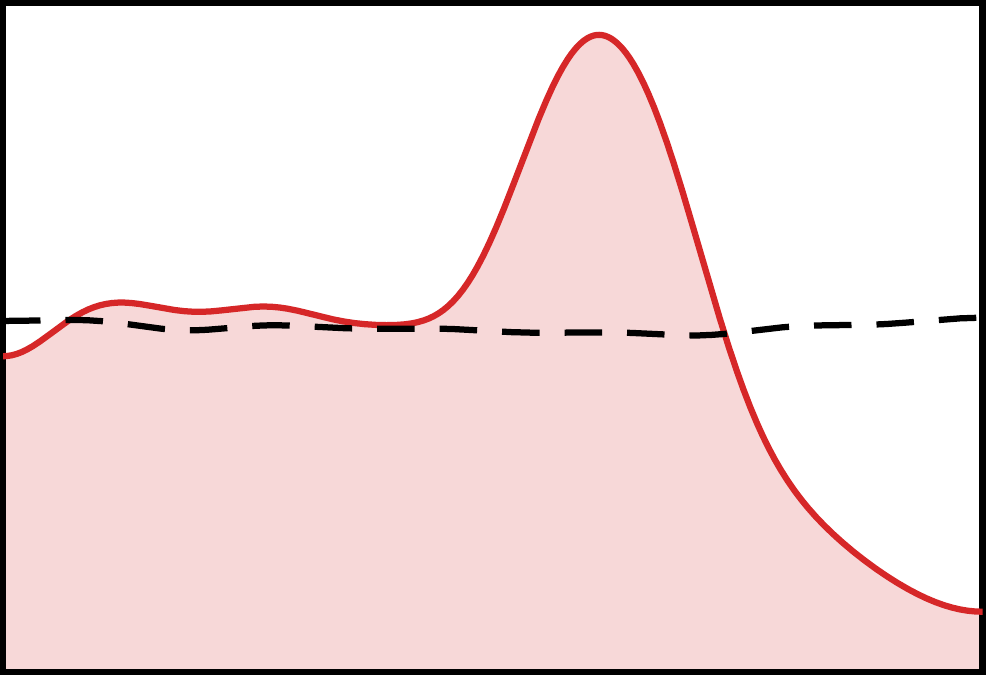}\end{minipage} & \begin{minipage}[t]{0.19\linewidth}\centering\hypminiVtwo{H3c-3}{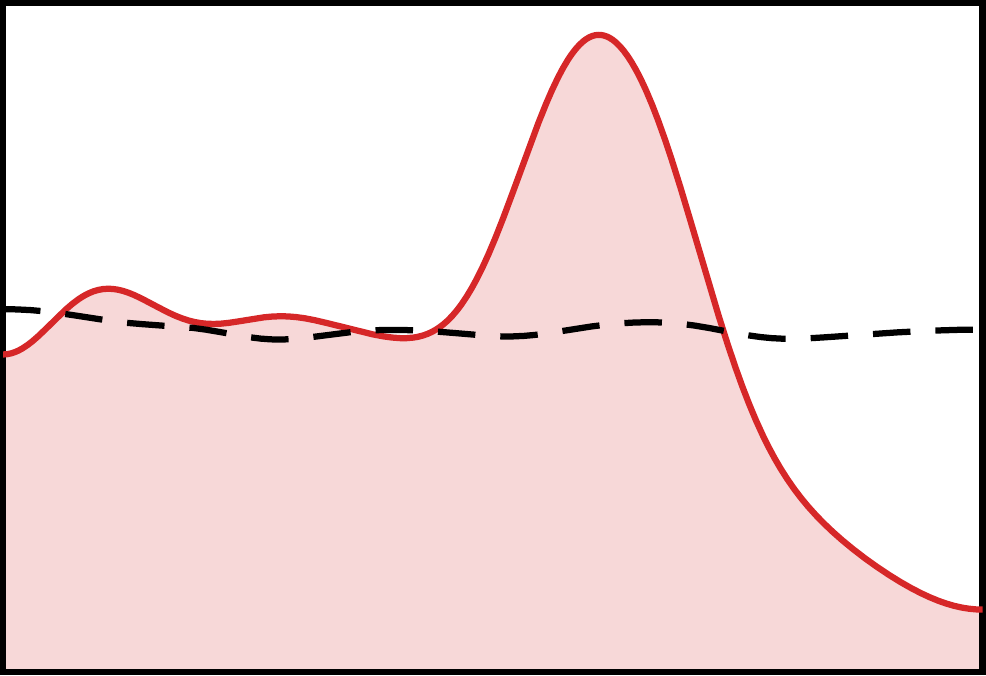}\end{minipage} & \begin{minipage}[t]{0.19\linewidth}\centering\hypminiVtwo{H4a}{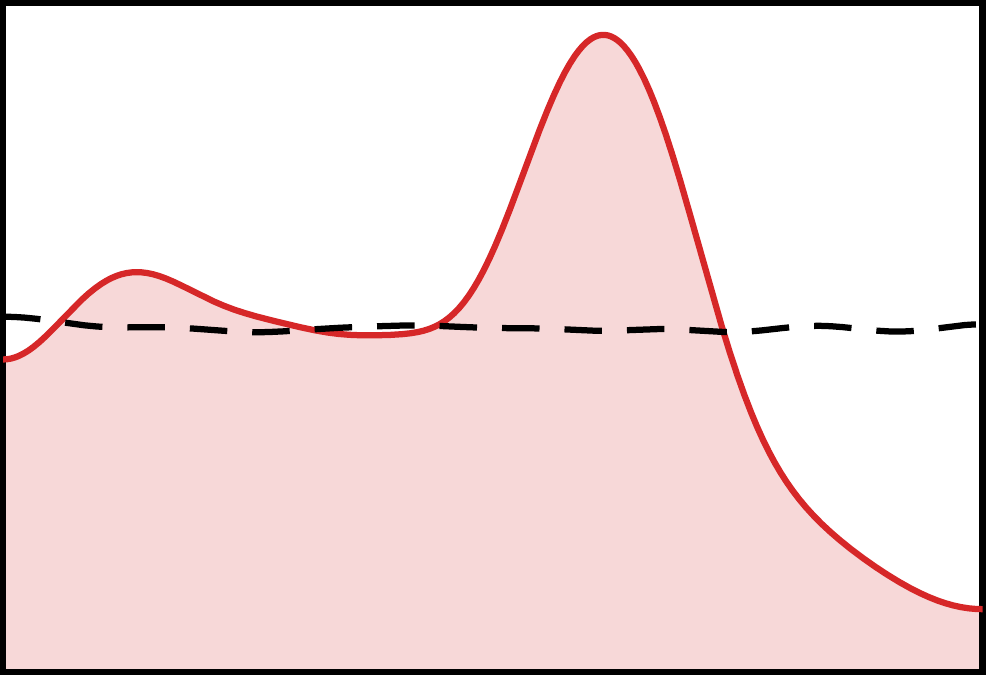}\end{minipage}\\[0.25em]
\begin{minipage}[t]{0.19\linewidth}\centering\hypminiVtwo{H5a}{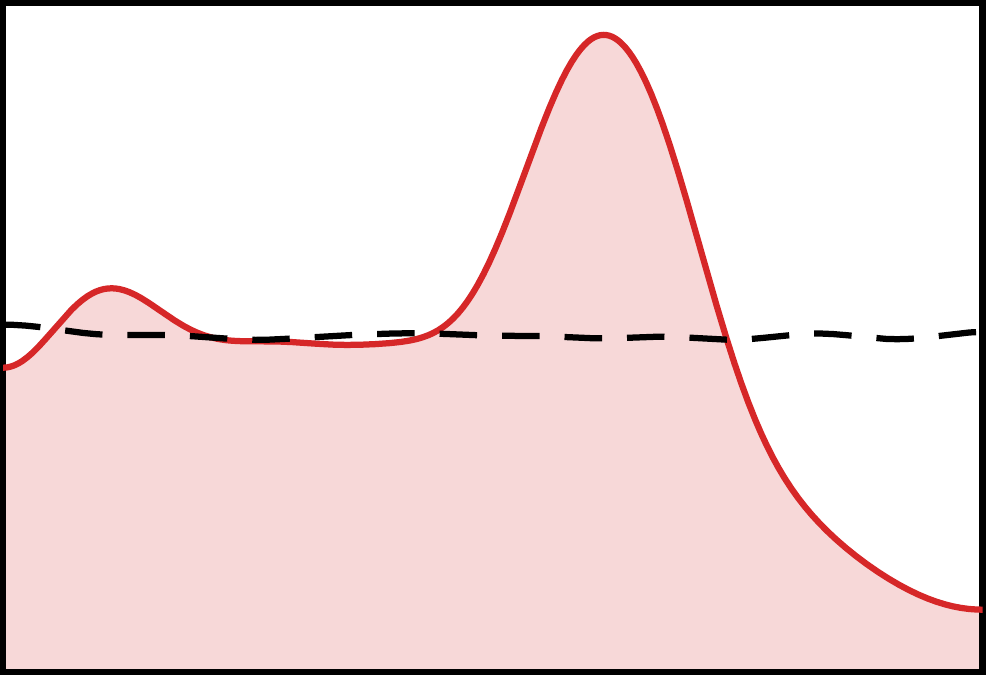}\end{minipage} & \begin{minipage}[t]{0.19\linewidth}\centering\hypminiVtwo{H5b}{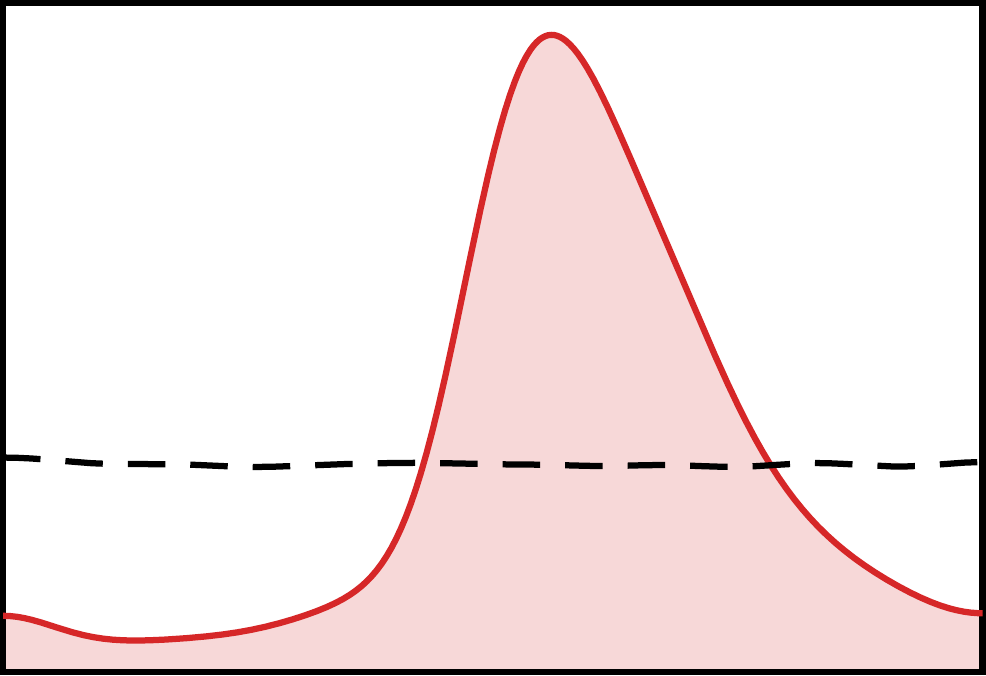}\end{minipage} & \begin{minipage}[t]{0.19\linewidth}\centering\hypminiVtwo{H6a}{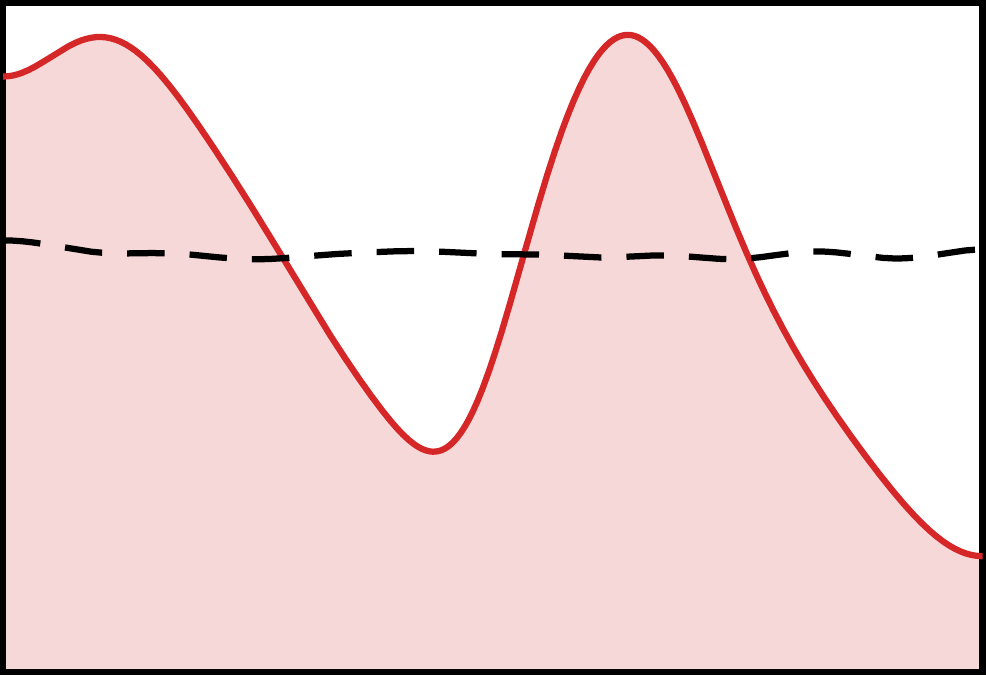}\end{minipage} & \begin{minipage}[t]{0.19\linewidth}\centering\hypminiVtwo{H7a}{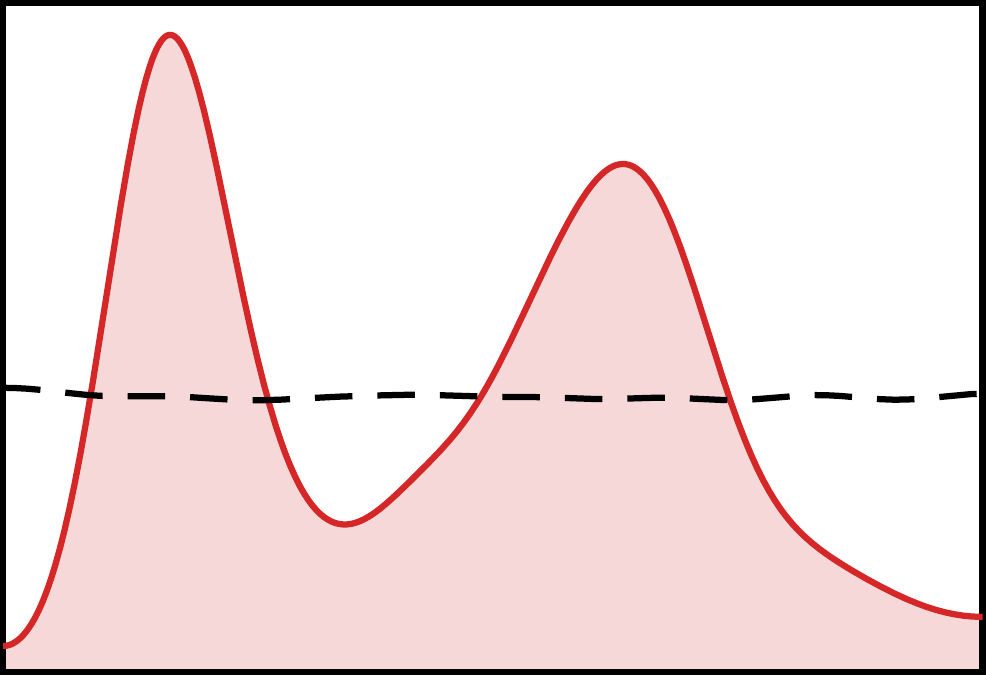}\end{minipage} & \begin{minipage}[t]{0.19\linewidth}\centering\hypminiVtwo{H8a}{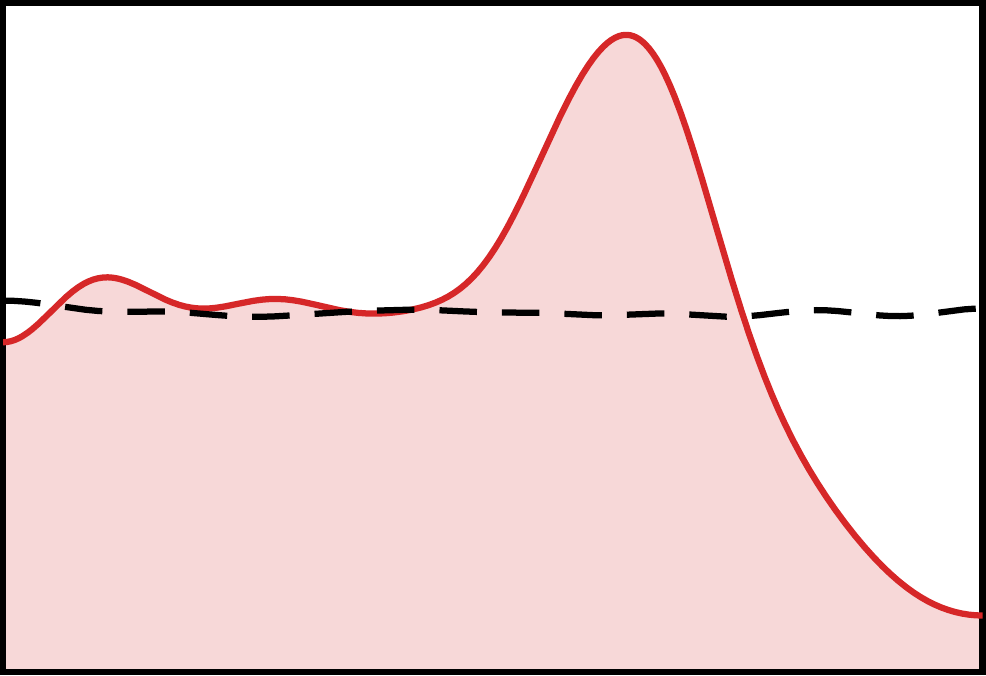}\end{minipage}\\[0.25em]
\begin{minipage}[t]{0.19\linewidth}\centering\hypminiVtwo{H8b}{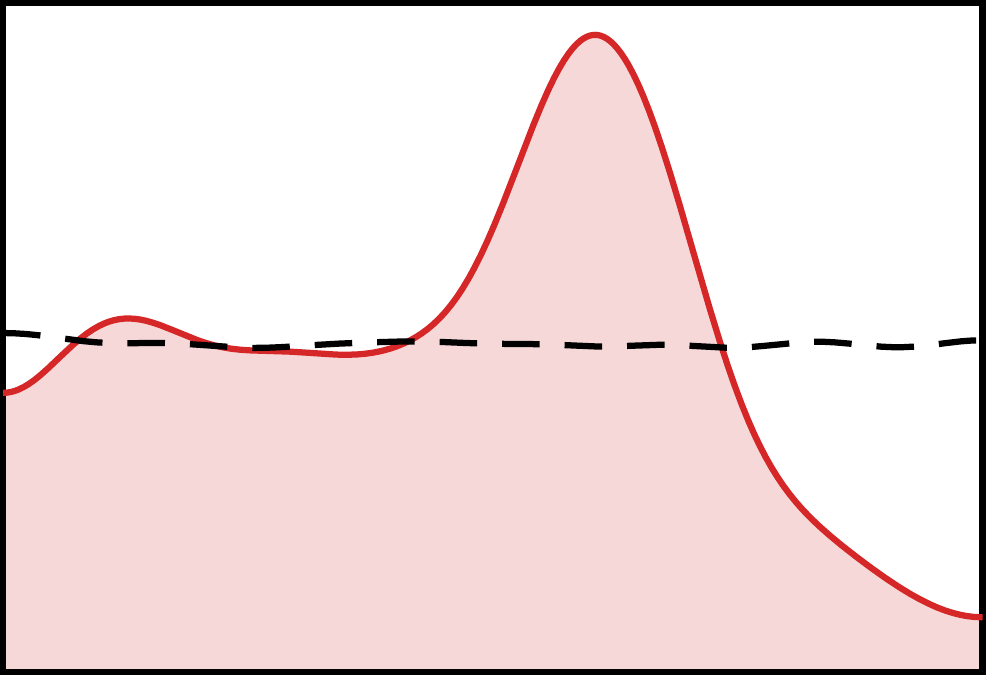}\end{minipage} & \begin{minipage}[t]{0.19\linewidth}\centering\hypminiVtwo{H9a}{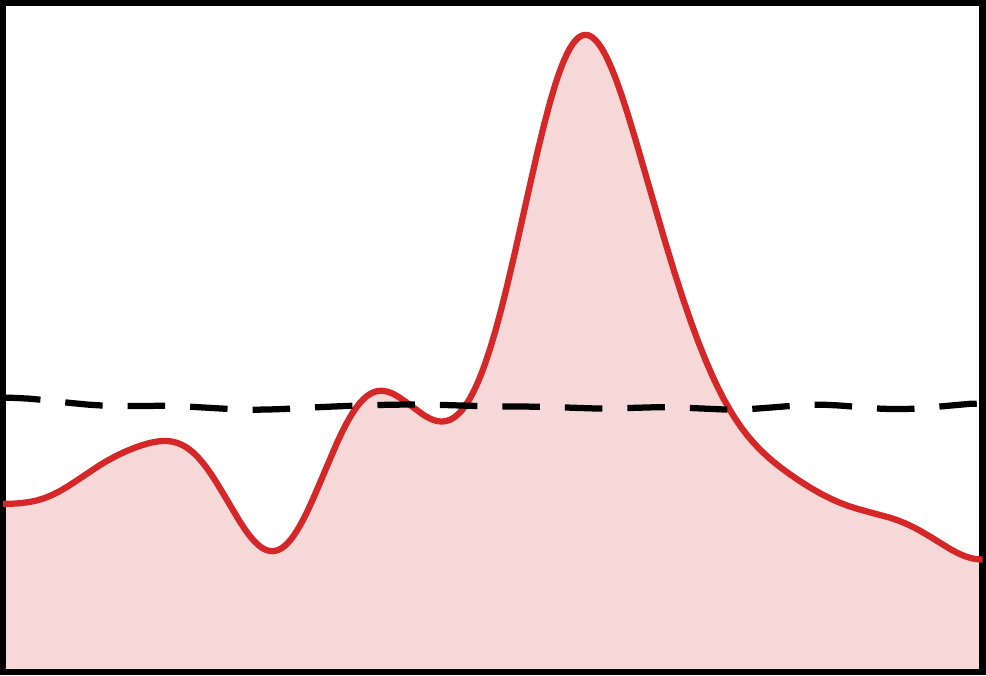}\end{minipage} & \begin{minipage}[t]{0.19\linewidth}\centering\hypminiVtwo{H9b}{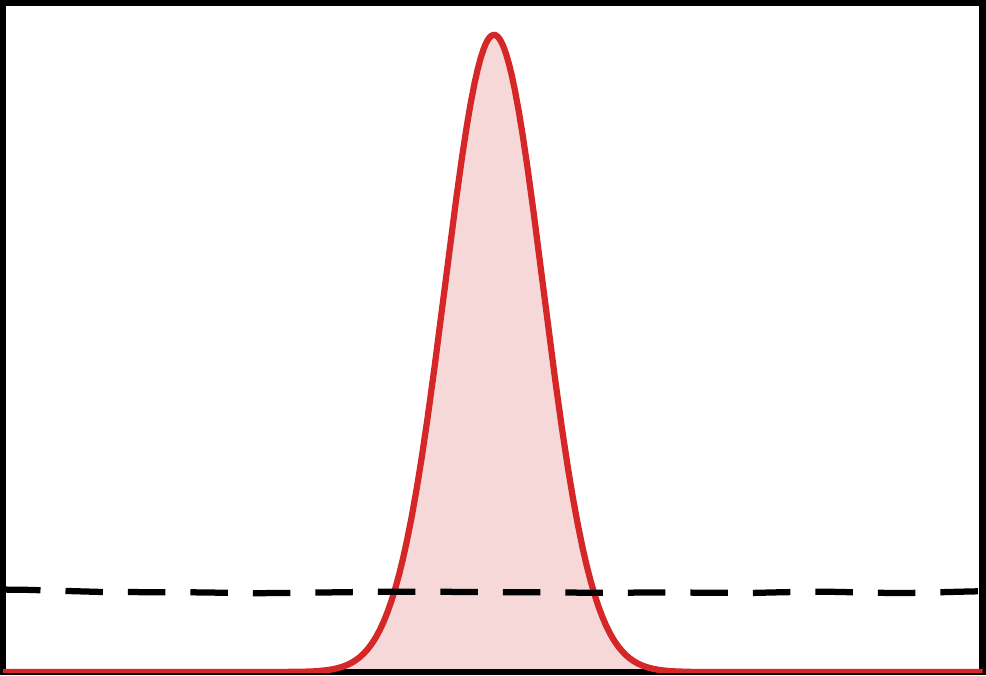}\end{minipage} & \begin{minipage}[t]{0.19\linewidth}\centering\hypminiVtwo{H10a}{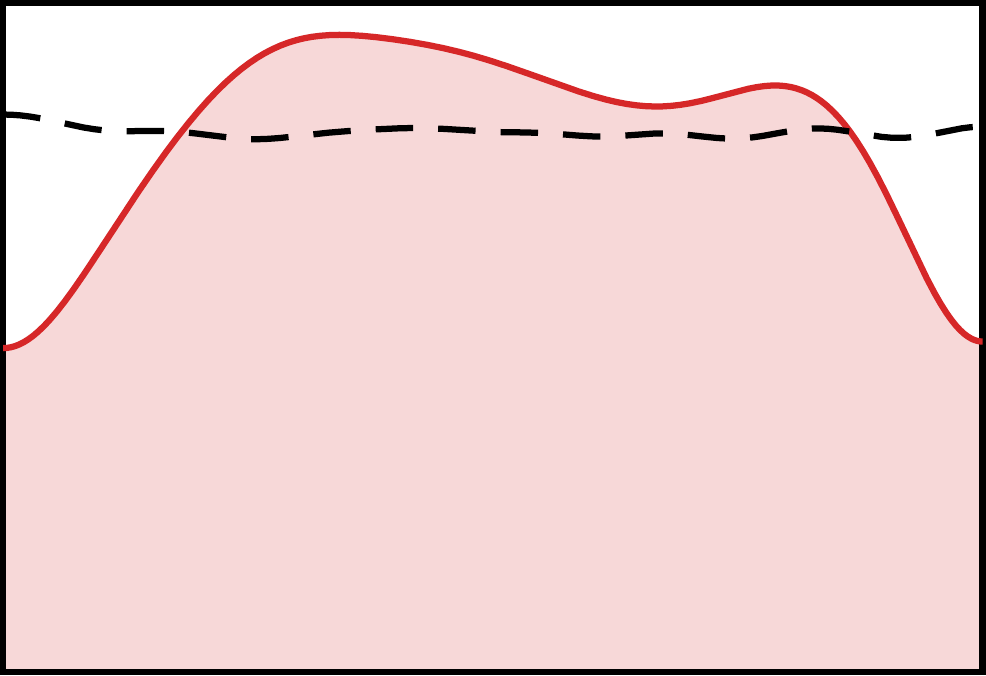}\end{minipage} & \begin{minipage}[t]{0.19\linewidth}\centering\hypminiVtwo{H10b}{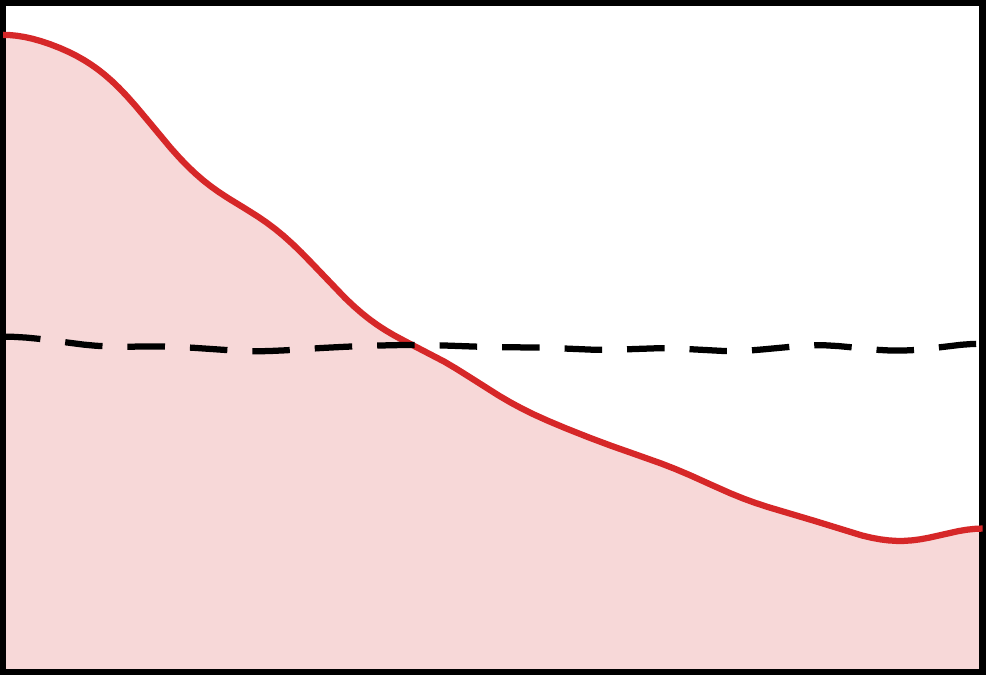}\end{minipage}\\[0.25em]
\end{tabular}
\end{center}

	\caption{\textbf{Alternative explanations do not remove physical misgeneralization.}
    }
    \label{fig:app:hypothesis-suite}
\end{figure}

\smallskip
\noindent\textbf{\xmark{} Hypothesis H1:} \emph{The rule for recovering the quantity creates the shift.}\par
\noindent To check whether the calculation used for recovery itself creates the shift, we hold the generated trajectories fixed and vary the summary of the posterior and the discrepancy used to recover $r$.
Specifically, we use the posterior mean in place of the posterior mode~(H1a), a discrepancy based on $\ell_1$ distance in place of the discrepancy based on squared error~(H1b), and both changes together~(H1c).
Yet, across all three variants, the distribution recovered from ground-truth trajectories remains close to the intended prior, while the distribution recovered from generated trajectories remains shifted~(\Tref{tab:app:hypothesis-v2}, H1a--H1c).
We therefore reject H1.

\smallskip
\noindent\textbf{\xmark{} Hypotheses H2--H3:} \emph{The model sees too little of the quantity mixture or trains too little to learn it.}\par
\noindent We next test whether the shift disappears after balancing coverage over the quantity or changing the amount of optimization.
To do so, we draw values of the quantity uniformly within {$256$ }bins before generating trajectories from the ground-truth map~(H2a), vary the number of epochs used for training {from $20$ to $6$~(H3a) or $60$~(H3b)}, and repeat the baseline across independent seeds~(H3c).
Sampling quantity values by strata still yields a shifted marginal after generation, changing the number of epochs modulates the severity without restoring the intended mixture, and independent seeds vary only slightly around the baseline value~(\Tref{tab:app:hypothesis-v2}, H2a--H3c).
We therefore reject H2--H3.

\smallskip
\noindent\textbf{\xmark{} Hypothesis H4:} \emph{The denoiser is too small.}\par
\noindent We test whether capacity is the bottleneck by training a wider U-Net {(base width $64$ instead of $32$) }while leaving the data, the rule for recovering the quantity, and the sampler fixed.
The recovered distribution remains substantially shifted~(\Tref{tab:app:hypothesis-v2}, H4a).
We therefore reject H4.

\smallskip
\noindent\textbf{\xmark{} Hypothesis H5:} \emph{The convolutional denoiser is too local.}\par
\noindent We then test the local structure of the denoiser by enlarging the convolutional kernel {from $3$ to $7$}~(H5a), and by replacing the convolutional denoiser with an MLP denoiser that can access the full trajectory at once~(H5b).
The {MLP has three hidden layers of width $1088$; we use a learning rate of $10^{-3}$ and a log-linear noise schedule spanning $[0.005,3]$.
The }larger convolutional kernel does not restore the intended mixture, and the MLP denoiser misgeneralizes even more strongly~(\Tref{tab:app:hypothesis-v2}, H5a--H5b).
We therefore reject H5.

\smallskip
\noindent\textbf{\xmark{} Hypotheses H6--H7:} \emph{The sampler or objective used for diffusion creates the shift.}\par
\noindent We isolate two choices in the diffusion implementation by {replacing $256$ DDPM steps with $50$ deterministic }DDIM steps~(H6a), and by replacing noise prediction with {{prediction of the clean trajectory} $x_0$}~(H7a).
Both changes leave the recovered distribution far from the intended prior~(\Tref{tab:app:hypothesis-v2}, H6a--H7a).
We therefore reject H6--H7.

\smallskip
\noindent\textbf{\xmark{} Hypothesis H8:} \emph{The length of the trajectory creates the shift.}\par
\noindent We test whether the horizon is responsible by using shorter {($H=32$, H8a) and longer ($H=128$, H8b) }trajectories drawn from the tent map {in place of $H=64$}.
Substantial drift remains at both horizons~(\Tref{tab:app:hypothesis-v2}, H8a--H8b), although the magnitude changes.
We therefore reject H8.

\smallskip
\noindent\textbf{\xmark{} Hypothesis H9:} \emph{Only diffusion models exhibit physical misgeneralization.}\par
\noindent We train an autoregressive Transformer and a VAE on trajectories drawn from the tent map.
{The Transformer has $4$ layers, width $216$, $6$ attention heads, and a vocabulary of $1024$ discretized states; the VAE has a one-dimensional latent variable and two hidden layers of width $1088$ for both the encoder and decoder.
}Both models induce shifted distributions over quantity values~(\Tref{tab:app:hypothesis-v2}, H9a--H9b); together with the MLP denoiser in H5b, this shows that the failure is not limited to diffusion.
We therefore reject H9.

\smallskip
\noindent\textbf{\cmark{} Hypothesis H10:} \emph{Changing the range of the quantity changes the strength of misgeneralization.}\par
\noindent We test which of a lower interval of $r$~($[0,0.5]$, H10a) or a higher interval~($[1.5,2]$, H10b) produces more drift.
We find that the lower interval yields a recovered distribution closer to the intended prior, while the higher interval produces larger drift~(\Tref{tab:app:hypothesis-v2}, H10a--H10b).
This suggests that the issue is mediated by differences in how data are arranged with respect to the quantity, consistent with the mechanism we identify in the main body.

Together, these ablation results provide strong evidence against the simpler explanations.

\clearpage

\section{Text-to-Speech Vignette}
\label{sec:app:tts-vignette}

In the main text, our focus was on physical sequence models; however, in a set of preliminary experiments, we also found a similar aggregate mismatch in the text-to-speech domain.
Here, data are waveforms rather than physical trajectories, and we recover speaking rate rather than path length, energy, or the parameter of a dynamical system.
Even so, the structural issue is familiar: the model is trained to synthesize audio from text, while training does not directly specify the distribution over how quickly the text is spoken.

For this investigation, we use the LJSpeech~\citep{Ito2017LJSpeech} dataset, and synthesize matched utterances with the SpeechBrain Tacotron~2 and HiFi-GAN pipeline~\citep{Shen2018Tacotron2,Kong2020HiFiGAN,Ravanelli2021SpeechBrain}. An spectrogram of a generated example utterance is shown in~\Fref[a]{fig:app:tts}.
For each real and generated utterance, we transcribe the waveform with wav2vec~2.0~\citep{baevski2020wav2vec}, count syllables with a deterministic vowel-group heuristic, and divide by waveform duration to recover speaking rate in syllables per second.
Since a mismatch in transcript content would make the rate comparison ambiguous, we only keep paired prompts for which the generated utterance has word error rate no worse than the real utterance for the same prompt.
Among the $777$ paired utterances that pass this filter, we recover a faster speaking-rate distribution from generated speech: the mean speaking rate is $4.356$ syllables/sec for generated speech versus $4.222$ for data, with $\TV=0.103$~(\Fref[b]{fig:app:tts}).

This vignette is not part of our predictive claims, but the direction of the mismatch is still mechanistically suggestive.
In particular, we find that the mel-spectrogram objective used in this pipeline allows for substantially larger speed-ups than slow-downs at equal mel loss~(\Fref[c]{fig:app:tts}).
To measure this asymmetry, we take each real utterance, apply a small voiced-frame slow-down, compute the induced mel-spectrogram loss, and then find the speed-up that gives the same loss.
Across utterances, the matched speed-ups are about $16.3\times$ larger than the corresponding slow-downs, with interquartile range $11.8$--$20.3$.
Thus, in a setting far from the physical systems studied above, we again find that an aggregate quantity changes in the direction that the objective penalizes comparatively weakly.

We stop short of making a full prediction, because text-to-speech composes multiple learned stages, and defining a composed kernel for this setup is a nontrivial task that we leave for future work.
In particular, Tacotron~2 first produces an acoustic representation, and HiFi-GAN then maps that representation into waveform space.
Since we measure speaking rate only after the waveform is produced, a full prediction for this setting would require a data deviation kernel that summarizes how errors of the acoustic model and the vocoder feed into one another.
We hope this example encourages follow-ups on how data deviation kernels can be composed across learned stages, so misgeneralization can be anticipated in domains well beyond the physical systems studied here.

\begin{figure}[H]
	\centering
        \captionsetup[subfigure]{oneside,margin={0.25cm,0cm}}
		\begin{subfigure}[t]{0.31\linewidth}
			\centering
			\maybeincludegraphics[height=1.35in]{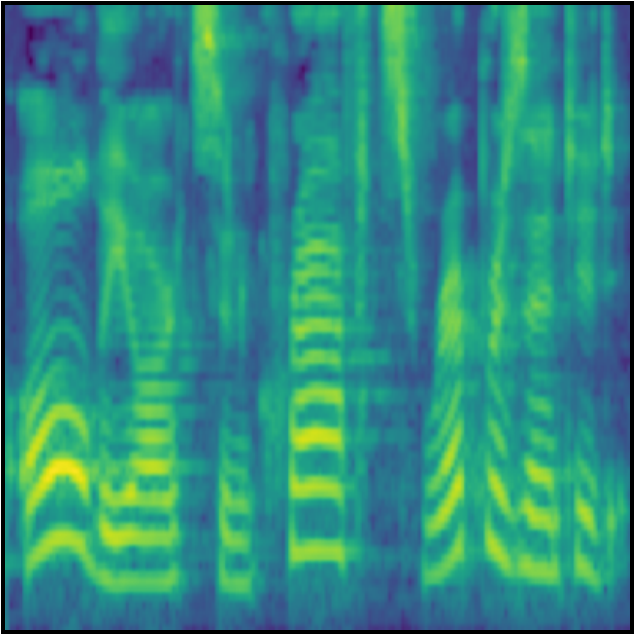}
			\caption{Example spectrogram of a model-generated sample.}
		\end{subfigure}\hfill
        \captionsetup[subfigure]{oneside,margin={0.8cm,0cm}}
		\begin{subfigure}[t]{0.31\linewidth}
			\centering
			\maybeincludegraphics[height=1.35in]{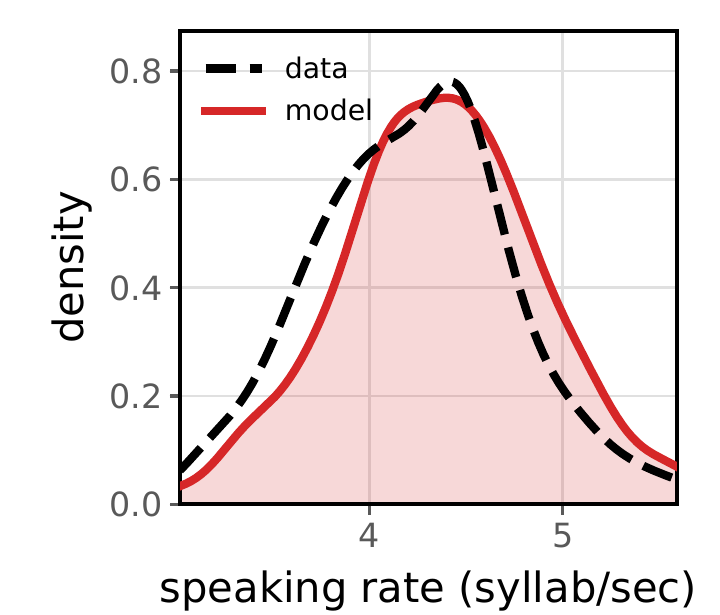}
			\caption{The model's upwards drift in speaking rate.}
		\end{subfigure}\hfill
        \captionsetup[subfigure]{oneside,margin={0.75cm,0cm}}
		\begin{subfigure}[t]{0.31\linewidth}
			\centering
			\maybeincludegraphics[height=1.35in]{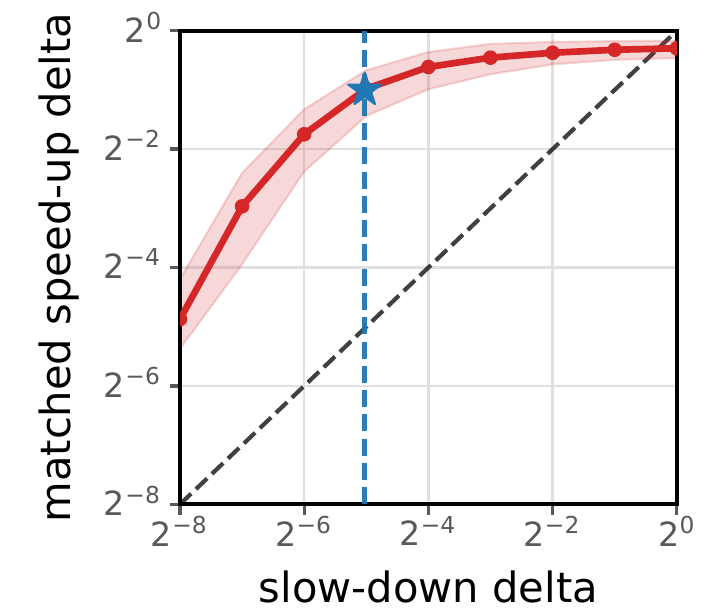}
            \caption{{Speed-ups \& slow-downs at equal mel loss.}}
		\end{subfigure}
		\caption{\textbf{Drift in speaking rate for generated speech.}
		When we compare real LJSpeech utterances to utterances synthesized from the same text with Tacotron~2 and HiFi-GAN, the generated utterances recover to a faster speaking-rate distribution than the training data implies.
		The time-warp probe shows that equal mel loss allows much larger speed-ups than slow-downs, pointing to the possibility that this is analogous to physical misgeneralization wherein intrinsic architectural quirks and low-cost directions under the loss get converted into a distribution shift through the physical measurement.}
	\label{fig:app:tts}
\end{figure}

\clearpage

\section*{NeurIPS Paper Checklist}

\begin{enumerate}

\item {\bf Claims}
    \item[] Question: Do the main claims made in the abstract and introduction accurately reflect the paper's contributions and scope?
    \item[] Answer: \answerYes{} 
    \item[] Justification: The abstract and introduction cover the physical misgeneralization phenomenon, the kernel mechanism, and the synthetic and applied experiments. These claims are developed in \Sref{sec:definition}--\Sref{sec:experiments} and scoped appropriately in \Sref{sec:limitations}. 
    \item[] Guidelines:
    \begin{itemize}
        \item The answer \answerNA{} means that the abstract and introduction do not include the claims made in the paper.
        \item The abstract and/or introduction should clearly state the claims made, including the contributions made in the paper and important assumptions and limitations. A \answerNo{} or \answerNA{} answer to this question will not be perceived well by the reviewers. 
        \item The claims made should match theoretical and experimental results, and reflect how much the results can be expected to generalize to other settings. 
        \item It is fine to include aspirational goals as motivation as long as it is clear that these goals are not attained by the paper. 
    \end{itemize}

\item {\bf Limitations}
    \item[] Question: Does the paper discuss the limitations of the work performed by the authors?
    \item[] Answer: \answerYes{} 
    \item[] Justification: We provide an extensive discussion of limitations in \Sref{sec:limitations} covering kernel specificity, generalizability beyond physical sequence models, and implications for fair representation. 
    \item[] Guidelines:
    \begin{itemize}
        \item The answer \answerNA{} means that the paper has no limitation while the answer \answerNo{} means that the paper has limitations, but those are not discussed in the paper. 
        \item The authors are encouraged to create a separate ``Limitations'' section in their paper.
        \item The paper should point out any strong assumptions and how robust the results are to violations of these assumptions (e.g., independence assumptions, noiseless settings, model well-specification, asymptotic approximations only holding locally). The authors should reflect on how these assumptions might be violated in practice and what the implications would be.
        \item The authors should reflect on the scope of the claims made, e.g., if the approach was only tested on a few datasets or with a few runs. In general, empirical results often depend on implicit assumptions, which should be articulated.
        \item The authors should reflect on the factors that influence the performance of the approach. For example, a facial recognition algorithm may perform poorly when image resolution is low or images are taken in low lighting. Or a speech-to-text system might not be used reliably to provide closed captions for online lectures because it fails to handle technical jargon.
        \item The authors should discuss the computational efficiency of the proposed algorithms and how they scale with dataset size.
        \item If applicable, the authors should discuss possible limitations of their approach to address problems of privacy and fairness.
        \item While the authors might fear that complete honesty about limitations might be used by reviewers as grounds for rejection, a worse outcome might be that reviewers discover limitations that aren't acknowledged in the paper. The authors should use their best judgment and recognize that individual actions in favor of transparency play an important role in developing norms that preserve the integrity of the community. Reviewers will be specifically instructed to not penalize honesty concerning limitations.
    \end{itemize}

\item {\bf Theory assumptions and proofs}
    \item[] Question: For each theoretical result, does the paper provide the full set of assumptions and a complete (and correct) proof?
    \item[] Answer: \answerYes{} 
    \item[] Justification: We provide formalizations in \Sref{sec:definition}--\Sref{sec:mechanism}. We also provide proofs for Lyapunov calculations of the synthetic families in \Aref{sec:app:lyapunov-sinusoid}--\Aref{sec:app:lyapunov-logistic}. 
    \item[] Guidelines:
    \begin{itemize}
        \item The answer \answerNA{} means that the paper does not include theoretical results. 
        \item All the theorems, formulas, and proofs in the paper should be numbered and cross-referenced.
        \item All assumptions should be clearly stated or referenced in the statement of any theorems.
        \item The proofs can either appear in the main paper or the supplemental material, but if they appear in the supplemental material, the authors are encouraged to provide a short proof sketch to provide intuition. 
        \item Inversely, any informal proof provided in the core of the paper should be complemented by formal proofs provided in appendix or supplemental material.
        \item Theorems and Lemmas that the proof relies upon should be properly referenced. 
    \end{itemize}

    \item {\bf Experimental result reproducibility}
    \item[] Question: Does the paper fully disclose all the information needed to reproduce the main experimental results of the paper to the extent that it affects the main claims and/or conclusions of the paper (regardless of whether the code and data are provided or not)?
    \item[] Answer: \answerYes{} 
    \item[] Justification: We describe the data generation, recovery rules, architectures, optimizer, sampling, and plotting conventions precisely in \Sref{sec:synthetic-suite}, \Sref{sec:experiments}, and \Aref{sec:appendix}. 
    \item[] Guidelines:
    \begin{itemize}
        \item The answer \answerNA{} means that the paper does not include experiments.
        \item If the paper includes experiments, a \answerNo{} answer to this question will not be perceived well by the reviewers: Making the paper reproducible is important, regardless of whether the code and data are provided or not.
        \item If the contribution is a dataset and\slash or model, the authors should describe the steps taken to make their results reproducible or verifiable. 
        \item Depending on the contribution, reproducibility can be accomplished in various ways. For example, if the contribution is a novel architecture, describing the architecture fully might suffice, or if the contribution is a specific model and empirical evaluation, it may be necessary to either make it possible for others to replicate the model with the same dataset, or provide access to the model. In general. releasing code and data is often one good way to accomplish this, but reproducibility can also be provided via detailed instructions for how to replicate the results, access to a hosted model (e.g., in the case of a large language model), releasing of a model checkpoint, or other means that are appropriate to the research performed.
        \item While NeurIPS does not require releasing code, the conference does require all submissions to provide some reasonable avenue for reproducibility, which may depend on the nature of the contribution. For example
        \begin{enumerate}
            \item If the contribution is primarily a new algorithm, the paper should make it clear how to reproduce that algorithm.
            \item If the contribution is primarily a new model architecture, the paper should describe the architecture clearly and fully.
            \item If the contribution is a new model (e.g., a large language model), then there should either be a way to access this model for reproducing the results or a way to reproduce the model (e.g., with an open-source dataset or instructions for how to construct the dataset).
            \item We recognize that reproducibility may be tricky in some cases, in which case authors are welcome to describe the particular way they provide for reproducibility. In the case of closed-source models, it may be that access to the model is limited in some way (e.g., to registered users), but it should be possible for other researchers to have some path to reproducing or verifying the results.
        \end{enumerate}
    \end{itemize}

\item {\bf Open access to data and code}
    \item[] Question: Does the paper provide open access to the data and code, with sufficient instructions to faithfully reproduce the main experimental results, as described in supplemental material?
    \item[] Answer: \answerYes{} 
    \item[] Justification: The repository includes scripts to download public inputs, generate synthetic data, train models, and render figures and tables.
    \item[] Guidelines:
    \begin{itemize}
        \item The answer \answerNA{} means that paper does not include experiments requiring code.
        \item Please see the NeurIPS code and data submission guidelines (\url{https://neurips.cc/public/guides/CodeSubmissionPolicy}) for more details.
        \item While we encourage the release of code and data, we understand that this might not be possible, so \answerNo{} is an acceptable answer. Papers cannot be rejected simply for not including code, unless this is central to the contribution (e.g., for a new open-source benchmark).
        \item The instructions should contain the exact command and environment needed to run to reproduce the results. See the NeurIPS code and data submission guidelines (\url{https://neurips.cc/public/guides/CodeSubmissionPolicy}) for more details.
        \item The authors should provide instructions on data access and preparation, including how to access the raw data, preprocessed data, intermediate data, and generated data, etc.
        \item The authors should provide scripts to reproduce all experimental results for the new proposed method and baselines. If only a subset of experiments are reproducible, they should state which ones are omitted from the script and why.
        \item At submission time, to preserve anonymity, the authors should release anonymized versions (if applicable).
        \item Providing as much information as possible in supplemental material (appended to the paper) is recommended, but including URLs to data and code is permitted.
    \end{itemize}

\item {\bf Experimental setting/details}
    \item[] Question: Does the paper specify all the training and test details (e.g., data splits, hyperparameters, how they were chosen, type of optimizer) necessary to understand the results?
    \item[] Answer: \answerYes{} 
    \item[] Justification: \Sref{sec:synthetic-suite} documents the experimental setup, and \Aref{sec:app:readout}--\Aref{sec:app:maze2d} give all details about normalization, optimizer, architecture, sampling, and quantity recovery.
    \item[] Guidelines:
    \begin{itemize}
        \item The answer \answerNA{} means that the paper does not include experiments.
        \item The experimental setting should be presented in the core of the paper to a level of detail that is necessary to appreciate the results and make sense of them.
        \item The full details can be provided either with the code, in appendix, or as supplemental material.
    \end{itemize}

\item {\bf Experiment statistical significance}
    \item[] Question: Does the paper report error bars suitably and correctly defined or other appropriate information about the statistical significance of the experiments?
    \item[] Answer: \answerYes{} 
    \item[] Justification: \Aref{sec:app:statistics} reports three seed values and paired significance tests for the main prediction and mitigation results.
    \item[] Guidelines:
    \begin{itemize}
        \item The answer \answerNA{} means that the paper does not include experiments.
        \item The authors should answer \answerYes{} if the results are accompanied by error bars, confidence intervals, or statistical significance tests, at least for the experiments that support the main claims of the paper.
        \item The factors of variability that the error bars are capturing should be clearly stated (for example, train/test split, initialization, random drawing of some parameter, or overall run with given experimental conditions).
        \item The method for calculating the error bars should be explained (closed form formula, call to a library function, bootstrap, etc.)
        \item The assumptions made should be given (e.g., Normally distributed errors).
        \item It should be clear whether the error bar is the standard deviation or the standard error of the mean.
        \item It is OK to report 1-sigma error bars, but one should state it. The authors should preferably report a 2-sigma error bar than state that they have a 96\% CI, if the hypothesis of Normality of errors is not verified.
        \item For asymmetric distributions, the authors should be careful not to show in tables or figures symmetric error bars that would yield results that are out of range (e.g., negative error rates).
        \item If error bars are reported in tables or plots, the authors should explain in the text how they were calculated and reference the corresponding figures or tables in the text.
    \end{itemize}

\item {\bf Experiments compute resources}
    \item[] Question: For each experiment, does the paper provide sufficient information on the computer resources (type of compute workers, memory, time of execution) needed to reproduce the experiments?
    \item[] Answer: \answerYes{} 
    \item[] Justification: \Aref{sec:appendix} reports the hardware used and approximate GPU-hour totals.
    \item[] Guidelines:
    \begin{itemize}
        \item The answer \answerNA{} means that the paper does not include experiments.
        \item The paper should indicate the type of compute workers CPU or GPU, internal cluster, or cloud provider, including relevant memory and storage.
        \item The paper should provide the amount of compute required for each of the individual experimental runs as well as estimate the total compute. 
        \item The paper should disclose whether the full research project required more compute than the experiments reported in the paper (e.g., preliminary or failed experiments that didn't make it into the paper). 
    \end{itemize}
    
\item {\bf Code of ethics}
    \item[] Question: Does the research conducted in the paper conform, in every respect, with the NeurIPS Code of Ethics \url{https://neurips.cc/public/EthicsGuidelines}?
    \item[] Answer: \answerYes{} 
    \item[] Justification: We use synthetic data, public benchmarks, and simulation-based experiments, with no data collected from human subjects. 
    \item[] Guidelines:
    \begin{itemize}
        \item The answer \answerNA{} means that the authors have not reviewed the NeurIPS Code of Ethics.
        \item If the authors answer \answerNo, they should explain the special circumstances that require a deviation from the Code of Ethics.
        \item The authors should make sure to preserve anonymity (e.g., if there is a special consideration due to laws or regulations in their jurisdiction).
    \end{itemize}

\item {\bf Broader impacts}
    \item[] Question: Does the paper discuss both potential positive societal impacts and negative societal impacts of the work performed?
    \item[] Answer: \answerYes{} 
    \item[] Justification: In \Sref{sec:limitations}, we engage in a nuanced discussion of the potential societal impacts that our work may have upon the sociotechnical realities of intelligent physical systems. We cover how this work may help make physical sequence models safer and more transparent, as well as how our mechanistic understanding relates to failures in fairness and bias in deep learning systems. 
    \item[] Guidelines:
    \begin{itemize}
        \item The answer \answerNA{} means that there is no societal impact of the work performed.
        \item If the authors answer \answerNA{} or \answerNo, they should explain why their work has no societal impact or why the paper does not address societal impact.
        \item Examples of negative societal impacts include potential malicious or unintended uses (e.g., disinformation, generating fake profiles, surveillance), fairness considerations (e.g., deployment of technologies that could make decisions that unfairly impact specific groups), privacy considerations, and security considerations.
        \item The conference expects that many papers will be foundational research and not tied to particular applications, let alone deployments. However, if there is a direct path to any negative applications, the authors should point it out. For example, it is legitimate to point out that an improvement in the quality of generative models could be used to generate Deepfakes for disinformation. On the other hand, it is not needed to point out that a generic algorithm for optimizing neural networks could enable people to train models that generate Deepfakes faster.
        \item The authors should consider possible harms that could arise when the technology is being used as intended and functioning correctly, harms that could arise when the technology is being used as intended but gives incorrect results, and harms following from (intentional or unintentional) misuse of the technology.
        \item If there are negative societal impacts, the authors could also discuss possible mitigation strategies (e.g., gated release of models, providing defenses in addition to attacks, mechanisms for monitoring misuse, mechanisms to monitor how a system learns from feedback over time, improving the efficiency and accessibility of ML).
    \end{itemize}
    
\item {\bf Safeguards}
    \item[] Question: Does the paper describe safeguards that have been put in place for responsible release of data or models that have a high risk for misuse (e.g., pre-trained language models, image generators, or scraped datasets)?
    \item[] Answer: \answerNA{} 
    \item[] Justification: We do not release high-risk pretrained artifacts. 
    \item[] Guidelines:
    \begin{itemize}
        \item The answer \answerNA{} means that the paper poses no such risks.
        \item Released models that have a high risk for misuse or dual-use should be released with necessary safeguards to allow for controlled use of the model, for example by requiring that users adhere to usage guidelines or restrictions to access the model or implementing safety filters. 
        \item Datasets that have been scraped from the Internet could pose safety risks. The authors should describe how they avoided releasing unsafe images.
        \item We recognize that providing effective safeguards is challenging, and many papers do not require this, but we encourage authors to take this into account and make a best faith effort.
    \end{itemize}

\item {\bf Licenses for existing assets}
    \item[] Question: Are the creators or original owners of assets (e.g., code, data, models), used in the paper, properly credited and are the license and terms of use explicitly mentioned and properly respected?
    \item[] Answer: \answerYes{} 
    \item[] Justification: Existing external assets such as data and models are cited in-text, and we follow corresponding licenses. All iconography is royalty-free. All other visual assets are original.
    \item[] Guidelines:
    \begin{itemize}
        \item The answer \answerNA{} means that the paper does not use existing assets.
        \item The authors should cite the original paper that produced the code package or dataset.
        \item The authors should state which version of the asset is used and, if possible, include a URL.
        \item The name of the license (e.g., CC-BY 4.0) should be included for each asset.
        \item For scraped data from a particular source (e.g., website), the copyright and terms of service of that source should be provided.
        \item If assets are released, the license, copyright information, and terms of use in the package should be provided. For popular datasets, \url{paperswithcode.com/datasets} has curated licenses for some datasets. Their licensing guide can help determine the license of a dataset.
        \item For existing datasets that are re-packaged, both the original license and the license of the derived asset (if it has changed) should be provided.
        \item If this information is not available online, the authors are encouraged to reach out to the asset's creators.
    \end{itemize}

\item {\bf New assets}
    \item[] Question: Are new assets introduced in the paper well documented and is the documentation provided alongside the assets?
    \item[] Answer: \answerYes{} 
    \item[] Justification: The repository documents the code and generated synthetic data procedures needed to reproduce the paper.
    \item[] Guidelines:
    \begin{itemize}
        \item The answer \answerNA{} means that the paper does not release new assets.
        \item Researchers should communicate the details of the dataset\slash code\slash model as part of their submissions via structured templates. This includes details about training, license, limitations, etc. 
        \item The paper should discuss whether and how consent was obtained from people whose asset is used.
        \item At submission time, remember to anonymize your assets (if applicable). You can either create an anonymized URL or include an anonymized zip file.
    \end{itemize}

\item {\bf Crowdsourcing and research with human subjects}
    \item[] Question: For crowdsourcing experiments and research with human subjects, does the paper include the full text of instructions given to participants and screenshots, if applicable, as well as details about compensation (if any)? 
    \item[] Answer: \answerNA{} 
    \item[] Justification: We do not involve crowdsourcing or experiments with human subjects. 
    \item[] Guidelines:
    \begin{itemize}
        \item The answer \answerNA{} means that the paper does not involve crowdsourcing nor research with human subjects.
        \item Including this information in the supplemental material is fine, but if the main contribution of the paper involves human subjects, then as much detail as possible should be included in the main paper. 
        \item According to the NeurIPS Code of Ethics, workers involved in data collection, curation, or other labor should be paid at least the minimum wage in the country of the data collector. 
    \end{itemize}

\item {\bf Institutional review board (IRB) approvals or equivalent for research with human subjects}
    \item[] Question: Does the paper describe potential risks incurred by study participants, whether such risks were disclosed to the subjects, and whether Institutional Review Board (IRB) approvals (or an equivalent approval/review based on the requirements of your country or institution) were obtained?
    \item[] Answer: \answerNA{} 
    \item[] Justification: This work does not involve crowdsourcing or human subjects, so IRB review is not applicable. 
    \item[] Guidelines:
    \begin{itemize}
        \item The answer \answerNA{} means that the paper does not involve crowdsourcing nor research with human subjects.
        \item Depending on the country in which research is conducted, IRB approval (or equivalent) may be required for any human subjects research. If you obtained IRB approval, you should clearly state this in the paper. 
        \item We recognize that the procedures for this may vary significantly between institutions and locations, and we expect authors to adhere to the NeurIPS Code of Ethics and the guidelines for their institution. 
        \item For initial submissions, do not include any information that would break anonymity (if applicable), such as the institution conducting the review.
    \end{itemize}

\item {\bf Declaration of LLM usage}
    \item[] Question: Does the paper describe the usage of LLMs if it is an important, original, or non-standard component of the core methods in this research? Note that if the LLM is used only for writing, editing, or formatting purposes and does \emph{not} impact the core methodology, scientific rigor, or originality of the research, declaration is not required.
    \item[] Answer: \answerNA{} 
    \item[] Justification: LLMs are not an important, original, or non-standard component of the core methodology.
    \item[] Guidelines:
    \begin{itemize}
        \item The answer \answerNA{} means that the core method development in this research does not involve LLMs as any important, original, or non-standard components.
        \item Please refer to our LLM policy in the NeurIPS handbook for what should or should not be described.
    \end{itemize}

\end{enumerate}

\end{document}